\documentclass[11pt]{article}

\usepackage[preprint]{acl}

\usepackage{times}
\usepackage{latexsym}

\usepackage[T1]{fontenc}

\usepackage[utf8]{inputenc}
\usepackage{makecell}

\usepackage{microtype}

\usepackage{inconsolata}

\usepackage{graphicx}
\usepackage{amsmath}
\usepackage{amssymb}
\usepackage{booktabs}
\usepackage{algorithm}
\usepackage{algorithmic}
\usepackage{tcolorbox}
\usepackage{multirow}
\tcbuselibrary{breakable}
\usepackage{xcolor}
\usepackage{float}     
\usepackage{adjustbox} 

\title{RICoT: Recurrent Interleaved Chain-of-Thought for \\ Page Stream Segmentation}

\author{
  Prashant Khatri\textsuperscript{1},
  Bhavana Malagavalli\textsuperscript{1},
  Saiful Haq\textsuperscript{2},
  Niyati Chhaya\textsuperscript{1} \\
  \textsuperscript{1}Hyperbots \quad
  \textsuperscript{2}Indian Institute of Technology Bombay
}

\begin{document}
\maketitle
\begin{abstract}
Large language models (LLMs) have renewed interest in Page Stream
Segmentation (PSS), splitting a multi-page file into its constituent
documents. Yet published LLM methods still frame the task as boundary
classification over one or a few neighboring pages, and commercial
document-splitting APIs do not disclose their methods. We reimagine
PSS for LLMs as a sequential process over the whole stream and
propose \textbf{RICoT} (Recurrent Interleaved Chain-of-Thought), a
prompting framework that works through the pages as a reader would:
within a sliding window it reasons about each page and commits its
boundary decision before moving on, and across windows it carries
forward a compact record of what it has read, keeping per-call cost
bounded. We also introduce \textbf{FinancePSS-205}, a benchmark of
$205$ real-world, de-identified financial page streams, to be released publicly. On
FinancePSS-205, RICoT matches or outperforms pairwise and windowed prompting
with the same backbone for each of three open-weight LLMs; on a TABME test subset, RICoT with GLM-5 reaches document F1
$0.847$ and boundary F1 $0.915$, above published zero-shot results. A
$2$B-parameter student, trained on a larger teacher's decisions with
and without the page-level reasoning, performs comparably in both
modes and comes within $0.05$ document F1 of the strongest prompted 
model on FinancePSS-205.
\end{abstract}

\section{Introduction}

\begin{figure}[t]
  \centering
  \includegraphics[width=\linewidth]{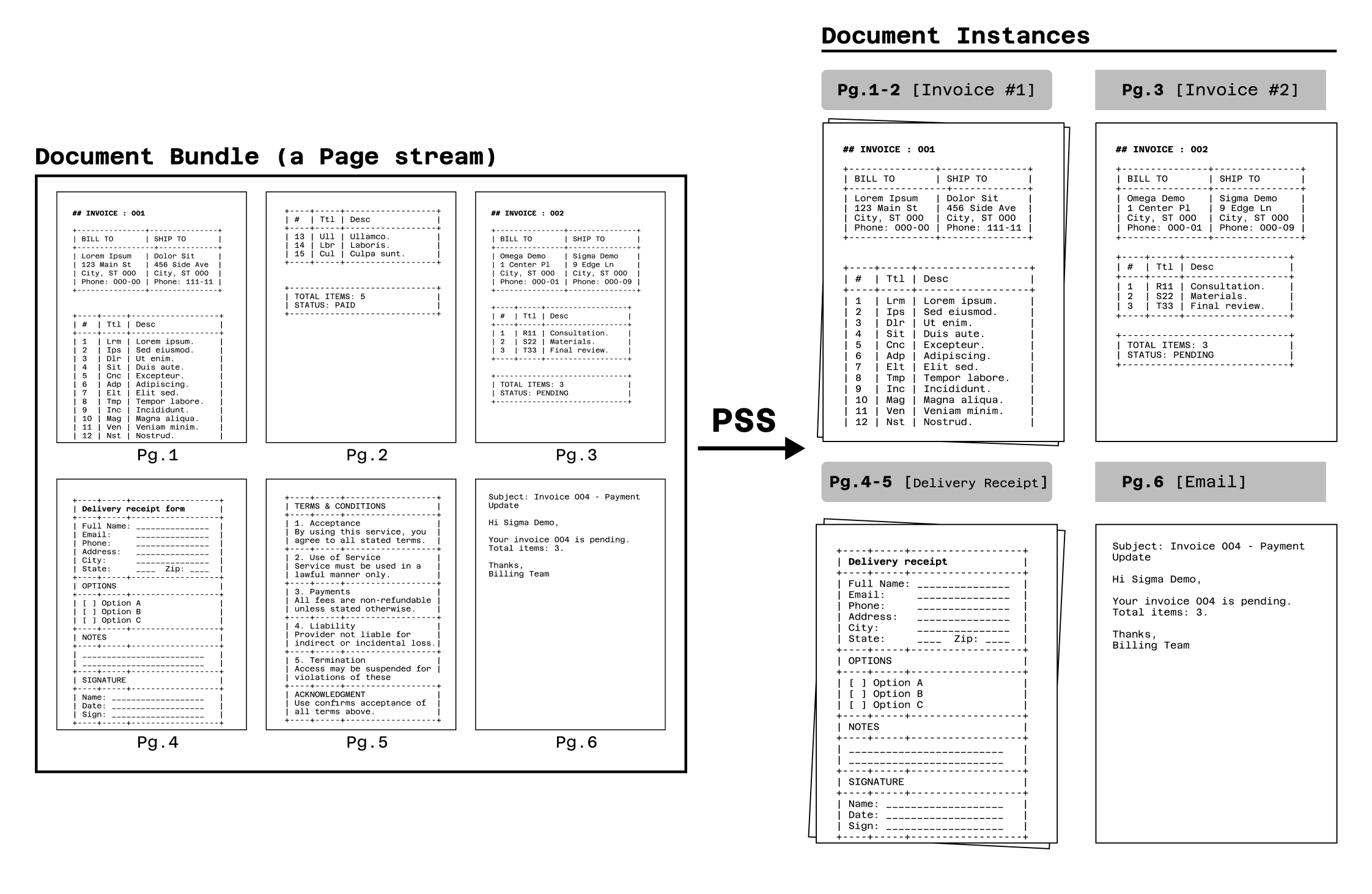}
  \caption{Page Stream Segmentation task. A stream of six pages is segmented into four documents: two invoices (pages 1--2 and 3), a delivery receipt (pages 4--5), and an email (page 6). Each document may span one or more consecutive pages.}
  \label{fig:pss_task}
\end{figure}

Document-level information retrieval typically assumes predefined document boundaries. However, real-world ``page streams''---sequences of scanned or digital pages---often bundle multiple documents into a single file. Page Stream Segmentation (PSS) partitions these streams into their constituent documents (Figure~\ref{fig:pss_task}), a task critical for finance automation (AP/AR), where batched invoices, purchase orders, and receipts are processed together; failing to segment them leads to merged field values, duplicated line items, and failed reconciliation.

Traditional PSS relies on \textbf{pairwise classification}~\cite{wiedemann2018,tabme2022,tabmepp}, deciding if page $P_i$ starts a new document based on $P_{i-1}$. This paradigm suffers from two primary limitations:
\begin{itemize}
    \item \textbf{Contextual Isolation:} Decisions often require long-range dependencies. For example, a remittance advice on page 11 may only be identifiable as part of a page 1 invoice via headers that a pairwise classifier cannot see.
    \item \textbf{Computational Redundancy:} Encoding each page twice---once as context and once as a target---doubles token costs without improving long-range reasoning.
\end{itemize}

Conversely, \textbf{full-stream processing} falters as sequence
length grows, since LLMs degrade in long contexts~\cite{liu2024lost}.
\textbf{Windowed classification}~\cite{pottanigari2026msbd} is a middle ground but is limited
by fixed window boundaries.

Our contributions are:

\begin{itemize}
\item \textbf{The RICoT framework} (Section~\ref{sec:ricot}): a
sliding-window PSS framework that combines interleaved per-page
chain-of-thought reasoning within each window with recurrent state
propagation across windows---a short reasoning history and an
active-segment summary that maintain document-level coherence for
arbitrarily long streams.

\item \textbf{A distilled $2$B student} (Section~\ref{sec:training}):
Qwen-3.5-2B trained on Kimi-K2-Instruct's RICoT outputs over a
finance-filtered subset of FinePDFs, with and without the page-level
reasoning; the two modes perform comparably, and the student comes
within $0.05$ DocF1 of the strongest prompted model on
FinancePSS-205.

\item \textbf{FinancePSS-205} (Section~\ref{sec:datasets}): a
new benchmark of $205$ real-world, de-identified AP/AR page
streams, to be released publicly with labels and evaluation scripts.\footnote{The dataset, evaluation code, prompt templates, and the additional materials referenced throughout this paper (``the public release'') are not yet public; a URL will be added in a future version of this preprint.}

\item \textbf{Extensive component ablations}
(Sections~\ref{sec:thinking-ablation} and~\ref{sec:results-ablations})
trace how each load-bearing element of RICoT---pagewise
interleaved reasoning, recurrent state propagation, the structured
decision schema, and the deployment-time \emph{expand}
post-processing---contributes to the framework's accuracy.
\end{itemize}

\section{Related Work}

\paragraph{Page stream segmentation.} PSS has progressed from
visual CNNs classifying adjacent page pairs~\cite{wiedemann2018}
through joint text-layout encoders~\cite{tabme2022} to
layout-feature embedding schemes~\cite{drumond2023layoutqt} and
efficient image-only pipelines~\cite{braz2021leveraging} on the
Tobacco800 benchmark~\cite{tobacco800,ltdl2007,agam2006cdip}.
\citet{tabmepp} adapt the paradigm to LLMs by prompting
instruction-tuned models on adjacent page pairs and obtain the
strongest zero-shot decoder baseline on TABME prior to our work;
\citet{heidenreich-etal-2025-page} extend the LLM treatment with
new metrics and stronger fine-tuned baselines. PSS has also been
formalized in adjacent media, e.g.\ multimodal Transformers on
comic-book page streams~\cite{serraortega2025cosmo}, and
real-world open benchmarks of Dutch government streams have
been released~\cite{wooir,openpss}.
A common thread across this prior work is the \emph{pairwise}
assumption: boundary decisions are made independently for each
page given only the immediately preceding page, which limits
context to the local pair. Concurrent MSBD~\cite{pottanigari2026msbd}
predicts all boundaries of a window in one call (the design of
our Windowed baseline) and degrades at larger windows; generation-based
segmenters~\cite{inan2022structured,li2026boundrl} are trained and
read the full input. RICoT is zero-shot, keeps windows bounded,
reasons per page, and carries state across windows. Commercial
document-AI services, such as LandingAI's Agentic Document
Extraction,\footnote{\url{https://docs.landing.ai/ade/ade-split}}
Reducto,\footnote{\url{https://docs.reducto.ai/api-reference/split}}
and LlamaIndex's LlamaParse,\footnote{\url{https://developers.llamaindex.ai/llamaparse/split/getting_started/};
all accessed October 2026.} also offer document splitting, but do
not publish their methods, so they cannot serve as reproducible
baselines.

\paragraph{Interleaved chain-of-thought reasoning.}
Chain-of-thought prompting~\cite{wei2022chain} elicits intermediate
reasoning steps before the final answer. The closest precedent for
the mechanism we use is ReAct~\cite{yao2023react}, which interleaves
\emph{thought} and \emph{action} steps for agentic tasks; RICoT
adapts the same interleaving principle to sequential
\emph{classification}, alternating per-page reasoning and per-page
decision within a window in one call, avoiding the per-page cost of
a multi-call agentic loop. Interleaved reasoning has recently been
explored as a training paradigm for LLMs~\cite{xie2026interleaved};
our contribution is at inference time, integrated with windowed
processing and recurrent state for multi-page sequential
classification.

\paragraph{Long-context state propagation.} Long-context use is
limited by the lost-in-the-middle effect~\cite{liu2024lost} and by
the cost of extending context windows. Architectural
mitigations~\cite{beltagy2020longformer,zaheer2020bigbird} and
recursive decomposition~\cite{zhang2025recursive} change the
underlying inference procedure; explicit memory-management
approaches such as MemGPT~\cite{packer2023memgpt} pass state across
turns through structured summaries. RICoT shares the
explicit-state philosophy, instantiated for PSS as a recurrent
active-segment summary plus a short reasoning history carried
across overlapping windows.

\section{RICoT Framework}
\label{sec:ricot}

\subsection{Task definition}
\label{sec:task-def}

Let $\mathbf{X}=(x_1,\dots,x_N)$ be a page stream of $N$ pages,
each represented by OCR-extracted text and layout features. PSS
maps $\mathbf{X}$ to binary boundary labels
$\mathbf{y}=(y_1,\dots,y_N)$ with $y_i=1$ iff page $x_i$ starts a
new document; by convention $y_1=1$. The labels induce a
partitioning into $K$ contiguous documents
$\{D_1,\dots,D_K\}$ where $D_k=(x_{s_k},\dots,x_{e_k})$ and
$\{s_1<\dots<s_K\}=\{i\mid y_i=1\}$.

\subsection{Sliding windows and the two-axis decision}
\label{sec:ricot-two-axis}

\begin{figure}[t]
  \centering
  \includegraphics[width=0.95\linewidth]{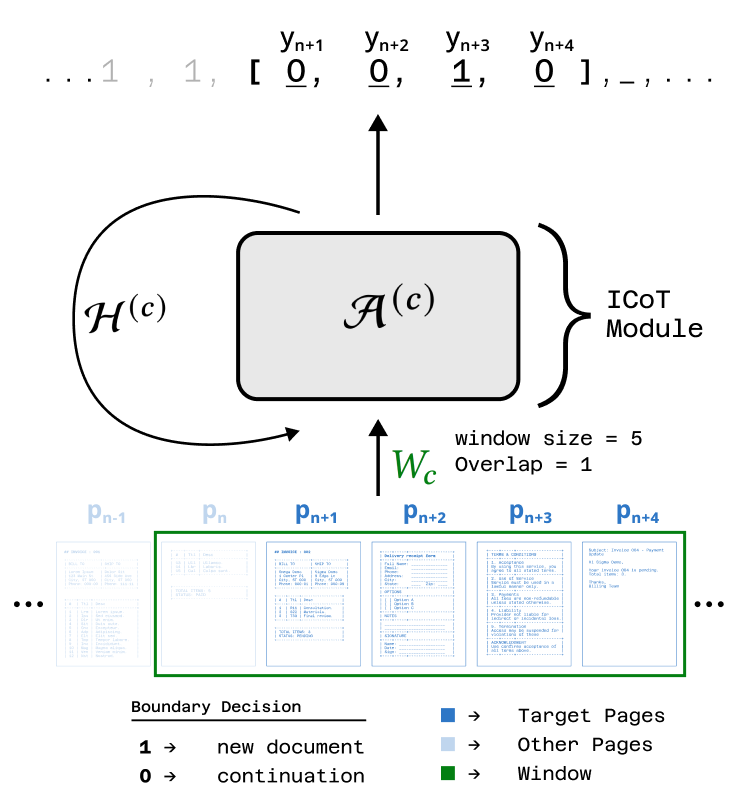}
  \caption{RICoT architecture. Window $W_c$ processes its pages
  with interleaved per-page reasoning and decisions, receiving
  reasoning history $\mathcal{H}^{(c)}$ and active-segment summary
  $\mathcal{A}^{(c)}$ from preceding windows. Recurrent state and
  the per-page two-axis decision (segment + sub-document) extend
  effective context beyond the window boundary while keeping
  per-window inference cost bounded.}
  \label{fig:ricot}
\end{figure}

RICoT (Figure~\ref{fig:ricot}) processes $\mathbf{X}$ with
\emph{token-budgeted} windows. Each window $W_c$ is packed greedily
with consecutive pages until the running token count would exceed a
budget $B$ or a page cap $w_{\max}$ is reached, with a one-page
overlap $o$ into the previous window.\footnote{We use $B=3600$
tokens, $w_{\max}=6$, $o=1$ across all experiments; values were
fixed \emph{a priori}, not tuned.} This adapts window size to page
density---short pages pack densely up to $w_{\max}$, long pages
truncate early---so context utilization is consistent across stream
geometries.

\paragraph{Two-axis decision.} For each page in the window the
model emits
\begin{equation}
  d_i \;=\; (y_i,\, z_i,\, \ell_i,\, g_i,\, c_i),
\end{equation}
where the \emph{segment axis} $y_i\in\{\mathrm{S},\mathrm{C}\}$
marks a new segment (S) or continuation (C); on continuations, the
\emph{sub-document axis} $z_i\in\{\mathrm{D},\varnothing\}$
optionally marks a nested document instance (e.g.\ an invoice inside
a statement of account). $\ell_i$ is a document-class label, $g_i$
a grounding fingerprint (page identifier + short evidence quote),
and $c_i$ a categorical confidence
($\textsc{high}{\mapsto}9$, $\textsc{mod}{\mapsto}6$,
$\textsc{low}{\mapsto}3$). $\ell_i$ is emitted only on S
decisions, so each segment has one label. Keeping segment-start and
sub-document-start as separate axes at decision time preserves both
signals for downstream control (Section~\ref{sec:ricot-expand}).

\subsection{Interleaved reasoning and recurrent state}
\label{sec:ricot-interleaved}

Within a window, RICoT generates per-page reasoning
(\texttt{<think>}) interleaved with the per-page decision
(\texttt{<decision>}) before moving to the next page. We refer to
this as \emph{pagewise thinking} (PwT); the alternative of emitting
one block of reasoning followed by all decisions is the
\emph{windowed thinking} variant we ablate in
Section~\ref{sec:thinking-ablation}. Interleaving grounds each
decision in evidence local to its target page.

Two recurrent state objects, $(\mathcal{H}^{(c)},\mathcal{A}^{(c)})$,
are propagated across windows. $\mathcal{H}^{(c)}$ is an anchored
recency window: up to the $k{=}6$ most recent thought--decision
pairs, always preserving the two decisions that opened the active
segment (joined by an elision marker once they age out).
$\mathcal{A}^{(c)}$ describes the currently open segment in two
parts: (i)~deterministic metadata
$(s,\ell_s,\textsc{pages-covered})$ derived by walking back to the
most recent S decision, and (ii)~a model-emitted
\texttt{<active\_segment\_carry\_forward>} block $\kappa^{(c)}$
summarizing anticipated content and progress for the next window.
The stream is wrapped with \texttt{<stream>}/\texttt{</stream>}
tags so the model suppresses $\kappa$ on the terminal window where
no successor exists. Each page is decided in the first window
containing it; later emissions for the overlap page are ignored.
A worked example is in Appendix~\ref{app:worked-example}.

\paragraph{Robust decoding.} A decision cache $\mathcal{C}$ stores
the most recent valid $d_i$ per page. After parsing, any missing,
malformed, or duplicate page is filled from $\mathcal{C}$ first; if
the page will recur in the next overlapping window the correction
is deferred; remaining gaps are repaired by a correction call, and
finally by rule-based fallbacks (continuation by default,
low-confidence \textsc{unknown} otherwise). The full procedure is
summarized in Algorithm~\ref{alg:ricot} (Appendix~\ref{app:algorithm}).

\subsection{Expand: deployment-time controllability}
\label{sec:ricot-expand}

Because the two-axis decision keeps sub-document starts as a
separate signal even when their parent segment is open, downstream
consumers can re-threshold what counts as a segment without
re-running inference. We use this for one specific operating
choice on FinancePSS-205, which we call \emph{expand}: when a
predicted sub-document of a priority class
(\texttt{invoice}, \texttt{statement}, \texttt{purchase\_order},
\texttt{packing\_slip}, \texttt{bill\_of\_lading},
\texttt{commercial\_invoice}, \texttt{packing\_list}) appears
inside a parent segment, the sub-document's start is promoted to a
true segment start in post-processing. The priority list is taken
from the document ontology of the AP/AR workflows the data come from
and was fixed before any FinancePSS evaluation; expand uses no
re-inference and no gold labels. We report its effect alongside the no-expand variant
throughout Section~\ref{sec:results} so its contribution is visible
rather than absorbed into the headline number.

\section{Distilling RICoT into a Small Model}
\label{sec:training}

A small model running the same framework is more attractive to deploy
than a large prompted one, so we distill RICoT into a $2$B
student: a larger teacher LLM runs RICoT over a training corpus,
and the student learns to produce the teacher's per-page decisions
both with and without the accompanying page-level reasoning.

\subsection{Training corpus from FinePDFs}
\label{sec:training-corpus}

We construct training data from FinePDFs~\cite{kydlicek2025finepdfs},
a publicly available PDF-sourced corpus
($\sim$$3$T tokens, $\sim$$475$M documents). We restrict to the
English subset ($\sim$$206$M documents) and isolate finance content
through a two-stage filter: Stage~1 is a high-recall keyword pass
over a tiered finance lexicon, and Stage~2 is an XGBoost
classifier over TF-IDF features trained on $\sim$$100$K Stage-1
documents labeled by an LLM judge for finance vs.\ non-finance. The
combined filter yields $12.3$M finance documents
($\sim$$100$B tokens). Lexicon tiers, classifier details, and
classification threshold tuning are in Appendix~\ref{app:finepdf}.

From this pool we sampled $300$K documents. Because FinePDFs lacks
page markers, we re-fetch each PDF via its Common Crawl URL and
re-OCR it per page, using FinePDFs' per-page language list to
verify page-count alignment and drop unsupported scripts; $150{,}448$
documents ($\sim$$50\%$) survive (Appendix~\ref{app:attrition}).

We then ran RICoT with the teacher model
(Kimi-K2-Instruct~\cite{kimi_k2}) over the $150{,}448$ documents,
producing $467{,}229$ chunk-level samples (each chunk is one
sliding-window span with the teacher's interleaved reasoning,
per-page decisions, and \texttt{active\_segment\_carry\_forward}
summary). Only $20{,}959$ ($13.9\%$) source documents contain two
or more segments; the remaining $86\%$ are self-contained
single-segment documents, reflecting FinePDFs' web-crawled
provenance.

\paragraph{Chunk selection.} We keep every multi-segment file and
every \emph{special} chunk in it (containing or immediately
preceding a segment start, or ending the file), subsampling other
chunks up to a cap. Single-segment files are downsampled within
embedding clusters of the teacher's free-form class labels, and
truncated or single-page documents are dropped. The final corpus
has $46{,}791$ documents ($20{,}959$ multi-segment, $25{,}832$
single-segment); caps will be included in the public release.

\subsection{Training with and without page-level reasoning}
\label{sec:training-modes}

Each training sample is a triple
$(\textsc{sys}, \textsc{ctx}, \textsc{out})$---system instruction,
stream context, chunk output---always rewritten together so the
schema declared, carried, and produced agree. With probability
$0.40$ per chunk, the sample is rendered without page-level
reasoning: the per-page \texttt{<think>} blocks are removed and
the instruction no longer mentions reasoning, so the target is the
teacher's decisions alone; otherwise it keeps the teacher's
interleaved reasoning and decisions. One student thus serves both
modes at deployment: with reasoning (Q3.5-2B$^{\star,\mathrm{PwT}}$)
and decisions only (Q3.5-2B$^{\star}$; Section~\ref{sec:models}).
A worked example of each will be included in the public release
(Appendix~\ref{app:samples}).

For operational convenience only, some samples also omit optional
decision fields (e.g., document class, confidence, grounding) or
the carry-forward \texttt{progress} field, so that one student
accepts reduced or full schemas without retraining; $y_i$ is never
removed, and we claim no accuracy contribution for this variation
(Appendix~\ref{app:sampling-config}).

\subsection{Training setup}
\label{sec:training-setup}

We fine-tune Qwen-3.5-2B~\cite{qwen} with QLoRA~\cite{qlora} of
rank~$32$ ($\alpha{=}64$, dropout $0$, target modules: all linear
projections) for $1$~epoch on the $46{,}791$-document training
corpus, with $\sim$$1{,}800$ samples held out for validation. We
use cosine learning-rate schedule
($8{\times}10^{-5}$, $8\%$ warmup), AdamW-$8$bit, bf16 with
gradient checkpointing, effective batch size $16$
(per-device $8$, grad-accumulation $2$). Training took
$\sim$$44$ hours on $4{\times}$A100 80GB. The full hyperparameter
set is in Appendix~\ref{app:hparams}. We report a single training
run; we did not sweep the sampling probabilities of
Section~\ref{sec:training-modes} or the LoRA rank.

\section{Experimental Setup}

\subsection{Datasets}
\label{sec:datasets}

We evaluate on three benchmarks spanning different domains and
complexity levels (Table~\ref{tab:datasets}).

\paragraph{TABME~\cite{tabme2022}.} A large synthetic page-stream
benchmark built from the Truth Tobacco Industry Documents archive,
in which $\sim$$44{,}769$ source PDFs were concatenated into
folders mimicking real folder length and type distributions. Each
source PDF forms one ground-truth segment, so TABME provides a
large, reproducible test of file-level segmentation. Inference cost
on the full $5{,}000$-stream test set is substantial, so we run
prompting experiments on a fixed random subset of $250$ streams
(\textbf{TABME-250}; stream IDs in Appendix~\ref{app:tabme-streams});
our distilled model is additionally evaluated on the full
$5{,}000$-stream test set, where numbers track the subset closely
(DocF1 $0.756$ vs.\ $0.770$ on $\star,\mathrm{PwT}$;
Section~\ref{sec:results-main}).

\paragraph{Tobacco800~\cite{tobacco800,ltdl2007,agam2006cdip}.} A
standard PSS benchmark from the same archive, containing scanned
letters, memos, and reports. Following~\citet{wiedemann_heyer}
and~\citet{tabme2022}, we evaluate on the last $259$ pages
from $153$ documents, ordered alphabetically by filename and
treated as a single continuous stream. It is near ceiling and
serves as a reference rather than discriminating benchmark.

\paragraph{FinancePSS-205 (new).} $205$ real-world page
streams from production AP/AR workflows, mixing invoices,
remittance advice, purchase orders, delivery receipts, statements,
and email, with typical finance bundling (invoice--remittance
pairs, PO--invoice matches, invoices nested in statements). Each
stream is one file exactly as received, so boundaries
between related same-vendor documents must be inferred from
content. A new document starts wherever a page begins a document
with its own transactional identity; labels were produced by the
research team (no inter-annotator agreement study; see Limitations). We will release the de-identified page-level,
spatially-aware text (the exact model input) with labels under CC BY-NC 4.0 (non-commercial research use) and
the evaluation scripts under Apache-2.0;
the dataset card and trivial baselines are
in Appendix~\ref{app:dataset-card}.

We use only the test partitions of the public datasets; no training
data from any benchmark is used in this work.

\begin{table}[t]
  \centering
  \scriptsize
  \setlength{\tabcolsep}{3pt}
    \begin{tabular}{@{}lccc@{}}
    \toprule
                       & TABME-250 & Tob800 & FinPSS \\
    \midrule
    Streams            & 250 & 1 & 205 \\
    Segments (docs)    & 2{,}627 (1{,}542) & 153 & 685 \\
    Pages              & 7{,}271 (4{,}240) & 259 & 1{,}772 \\
    Pages/stream       & 29.1 / 27 / 3 / 81 & 259$^\dagger$ & 8.6 / 5 / 2 / 82 \\
    Segs/stream        & 10.5 / 10 / 3 / 19 & 153$^\dagger$ & 3.3 / 2 / 1 / 44 \\
    Segment length     & 2.8 / 1 / 1 / 20 & 1.7 / 1 / 1 / 17 & 2.6 / 2 / 1 / 66 \\
    \bottomrule
  \end{tabular}
\caption{Test-set statistics. TABME segment/page counts shown
  as \emph{total (unique)} since TABME composes its streams from a
  shared pool of source files (Appendix~\ref{app:uncertainty}).
  Stream statistics reported as avg/med/min/max;
  $^\dagger$~Tobacco800 is evaluated as a single continuous stream,
  so per-stream distributional statistics do not apply.}
  \label{tab:datasets}
\end{table}

\subsection{Models}
\label{sec:models}

\paragraph{Prompted LLMs.}
\textbf{GLM-4.7-flash}~\cite{glm} is the $\sim$$30$B
small variant of the GLM-4.7 family, our small-model prompting
baseline---roughly $15\times$ the size of our distilled student;
\textbf{GLM-4.7}~\cite{glm} is the full GLM-4.7 model;
\textbf{GLM-5}~\cite{glm5} is a $744$B-parameter
Mixture-of-Experts model released under the MIT license by Z.ai;
\textbf{Claude Haiku 4.5}~\cite{claude_haiku45} is included as a
non-GLM frontier prompted-model point of comparison. By default we
run RICoT with each model's internal extended thinking
\emph{off} (Section~\ref{sec:thinking-ablation}).

\paragraph{Distilled student.} \textbf{Qwen-3.5-2B}~\cite{qwen}
distilled from Kimi-K2-Instruct~\cite{kimi_k2} as described in
Section~\ref{sec:training}; the base model prompted with RICoT
\emph{without} fine-tuning is reported as a control.

\paragraph{Table labels.}
\textbf{G4.7-F} for GLM-4.7-flash,
\textbf{G4.7} for GLM-4.7,
\textbf{G5} for GLM-5,
\textbf{C-H4.5} for Claude Haiku 4.5, and
\textbf{Q3.5-2B} for the Qwen-3.5-2B base model. The
distilled-student variants are denoted
\textbf{Q3.5-2B$^\star$} (no pagewise thinking) and
\textbf{Q3.5-2B$^{\star,\mathrm{PwT}}$} (with pagewise thinking).

\subsection{Evaluation metrics}
\label{sec:metrics}

We report five metrics spanning page, segment, and stream
granularities. \textbf{PageF1}~\cite{heidenreich-etal-2025-page,tabmepp}
is page-level boundary-classification F1. \textbf{B-F1}~\cite{pottanigari2026msbd}
is the same F1 over adjacent-page decisions only, excluding each
stream's first page. \textbf{DocF1}
\cite{tabmepp,heidenreich-etal-2025-page} is F1 over documents,
each represented as the tuple of its page indices; a true positive
is an exact tuple match. All F1 scores are micro-averaged over
streams.
$\mathrm{\textbf{MNDD}}_\mu$
(per-stream mean of Minimum Number of Drag-and-Drops) is a
user-centric metric counting the minimum drag-and-drop operations
needed to reconcile a prediction with the ground truth in a
folder-management UI; lower is better. \textbf{STP\%} (Straight-Through
Pass) is the percentage of streams with zero boundary errors.
All metrics are computed on the segment axis $y$ only and ignore
class labels; the sub-document axis enters only through the
with-expand rows (Section~\ref{sec:ricot-expand}).

\subsection{Baselines}
\label{sec:baselines}

Baselines are re-run on our infrastructure with the same
backbones as RICoT.

\textbf{Pairwise.} Each adjacent page pair $(p_{i-1},p_i)$ is
classified independently for whether $p_i$ starts a new
document~\cite{tabmepp,heidenreich-etal-2025-page}; window size
$w{=}2$, overlap $o{=}1$. Internal extended thinking enabled
(its strongest configuration per Section~\ref{sec:thinking-ablation}).

\textbf{Windowed.} A fixed-size window of $w{=}5$ pages is shown
to the LLM with a single decision prompt asking for boundary
labels across the window; windows overlap by $o{=}1$. Internal
extended thinking enabled. The same single-call, multi-boundary
design was proposed concurrently as MSBD~\cite{pottanigari2026msbd}.

\subsection{Implementation details}
\label{sec:impl}

The RICoT sliding window is sized by a constant token budget
($B{=}3600$), a per-window page cap ($w_{\max}{=}6$), and a
one-page overlap ($o{=}1$); these were fixed \emph{a priori} and
held constant across all experiments (sweep in
Appendix~\ref{app:window-sweep}). Pages whose tokenized length
exceeds a fixed per-page threshold are truncated using a
head-and-tail rule (first $500$ + last $300$ tokens, with an
explicit truncation marker between), since segmentation signal
concentrates at document openings and closings. All inference uses
temperature $0.001$ (top-$p$ $1$) for both prompting and the
distilled model. The inference configuration is described in
Appendix~\ref{app:prompts}; structural prompt templates will be included
in the public release.

Prompted results are single runs; distilled-model FinancePSS and
Tobacco800 numbers are means of $5$ inference runs. Run-to-run
variability and confidence intervals are in
Appendix~\ref{app:uncertainty}.

\section{Results}
\label{sec:results}

FinancePSS cells read
\emph{with-expand (no-expand)}, e.g.\ $0.832\,(0.795)$;
superscripts $\mathrm{iT}$, $\mathrm{PwT}$, $\star$ denote
internal-thinking, pagewise-thinking, and fine-tuned runs, and
em-dashes mark cells we did not run.

\subsection{Headline cross-dataset results}
\label{sec:results-main}

Table~\ref{tab:main-results} reports RICoT across all four prompted
models and two distilled-model variants on the three benchmarks.
The strongest prompted model (GLM-5) reaches DocF1 $0.967$ on
Tobacco800 and $0.847$ on TABME-250, and the strongest
prompted-model FinancePSS configuration (GLM-4.7) reaches $0.832$
with expand. The distilled $2$B model
($\text{Qwen-3.5-2B}^{\star,\mathrm{PwT}}$) reaches $0.785$ on
FinancePSS-205 and $0.770$ on the full $5{,}000$-stream TABME test
set without exposure to TABME training data. Its gap to the
strongest prompted model is $0.047$ DocF1 on FinancePSS-205,
$0.040$ on Tobacco800, and $0.091$ on TABME-250 (same streams). The same base
Qwen-3.5-2B prompted with RICoT but \emph{without} fine-tuning
reaches only $0.445$ / $0.396$ / $0.281$ on the three datasets,
isolating the contribution of distillation from
base-model capacity and framework prompts. The two student modes
perform comparably ($0.785$ vs.\ $0.773$ on FinancePSS-205, with
overlapping intervals in Appendix~\ref{app:uncertainty}; $0.770$
vs.\ $0.774$ on full TABME), so page-level reasoning can be skipped
at inference.

\begin{table*}[t]
  \centering
  \scriptsize
  \setlength{\tabcolsep}{3.5pt}
    \begin{tabular}{lccccccccc}
    \toprule
    & \multicolumn{3}{c}{\textbf{FinancePSS-205}} & \multicolumn{3}{c}{\textbf{Tobacco800}} & \multicolumn{3}{c}{\textbf{TABME-250}} \\
    \cmidrule(lr){2-4}\cmidrule(lr){5-7}\cmidrule(lr){8-10}
    Model & DocF1 & PageF1 & $\mathrm{MNDD}_\mu$ & DocF1 & PageF1 & $\mathrm{MNDD}_\mu$ & DocF1 & PageF1 & $\mathrm{MNDD}_\mu$ \\
    \midrule
    G4.7-F  & 0.762\,(0.741) & 0.898\,(0.884) & 1.005\,(1.225) & 0.867 & 0.949 & 48.0 & 0.683 & 0.829 & 5.65 \\
    G4.7    & \textbf{0.832}\,(0.795) & \textbf{0.916}\,(0.884) & \textbf{0.848}\,(1.255) & 0.961 & \textbf{0.987} & 7.0 & 0.794 & 0.892 & 3.31 \\
    C-H4.5  & 0.776\,(0.739) & 0.880\,(0.831) & 1.172\,(1.647) & 0.957 & 0.984 & 8.0 & 0.816 & 0.910 & 2.88 \\
    G5      & 0.826\,(0.769) & 0.906\,(0.864) & 0.975\,(1.422) & \textbf{0.967} & \textbf{0.987} & \textbf{3.0} & \textbf{0.847} & \textbf{0.922} & \textbf{2.40} \\
    \midrule
    Q3.5-2B (base) & 0.445 & 0.688 & 3.103 & 0.396 & 0.664 & 167.0 & 0.281 & 0.563 & 14.62 \\
    Q3.5-2B$^{\star}$           & 0.773\,(0.763) & 0.904\,(0.892) & 0.897\,(1.079) & 0.914 & 0.954 & 26.6 & 0.753 / 0.774$^{\mathrm{F}}$ & 0.873 / 0.894$^{\mathrm{F}}$ & 4.56 / 4.21$^{\mathrm{F}}$ \\
    Q3.5-2B$^{\star,\mathrm{PwT}}$ & 0.785\,(0.779) & 0.901\,(0.897) & 1.044\,(1.073) & 0.927 & 0.965 & 18.6 & 0.756 / 0.770$^{\mathrm{F}}$ & 0.885 / 0.896$^{\mathrm{F}}$ & 4.20 / 4.30$^{\mathrm{F}}$ \\
    \bottomrule
  \end{tabular}
\caption{Headline cross-dataset results for RICoT. DocF1 / PageF1
  higher is better; $\mathrm{MNDD}_\mu$ lower. FinancePSS numbers
  show with-expand outside parentheses and without-expand inside;
  without expand, predicted sub-document starts count as
  continuations, which is how TABME and Tobacco800 are scored
  throughout. TABME-250 / TABME-full reported separately for the
  distilled model.
  Bold marks the column best.
  $^\star$ fine-tuned rank-$32$ QLoRA;
  $^{\mathrm{PwT}}$ pagewise thinking enabled.
  $^{\mathrm{F}}$ Full $5{,}000$-stream TABME test set
  (vs.\ $250$-stream prompting subset).}
  \label{tab:main-results}
\end{table*}

\subsection{Comparison to prior published TABME baselines}
\label{sec:results-prior}

Table~\ref{tab:prior-tabme} compares RICoT with published TABME
results~\citep{tabmepp,heidenreich-etal-2025-page,pottanigari2026msbd}.
These span different test sets and OCR versions (table footnotes)
and are indicative only.
RICoT with GLM-5 (DocF1 $0.847$) beats the strongest published
zero-shot decoder (GPT-4o-ZS) by $+0.144$ DocF1 and $+32$ STP
points; the distilled $2$B model on the full test set ($0.770$)
beats the XGBoost and DiT baselines of~\citet{tabmepp} and GPT-4o-ZS
by $+0.067$. On boundary F1, we list for each model of~\citet{pottanigari2026msbd}
its best result across all their methods (PC, BD, MSBD), window
sizes, and inputs: $0.866$ (Claude Haiku~4.5), $0.887$ (Gemini 3.1
Flash-Lite), and $0.876$ (GPT-5.4-mini). RICoT, using text only,
exceeds all of them: GLM-5 reaches $0.915$ and Claude Haiku~4.5
$0.901$ ($+0.035$ over the same model's best), and even the $2$B
students ($0.890$, $0.887$) match or exceed them. None of our models approaches the fine-tuned LLMs (Phi-3-mini,
Mistral-7B) trained on TABME's training split; the gap reflects both TABME's file-level
segment granularity and genuine errors of our models
(Section~\ref{sec:analysis}).

\begin{table}[t]
  \centering
  \small
  \setlength{\tabcolsep}{4pt}
    \begin{adjustbox}{max width=\columnwidth}
    \begin{tabular}{lllrrrr}
    \toprule
    Model & Type & Input & DocF1 & B-F1 & STP\% & MNDD \\
    \midrule
    XGBoost     & \textsc{ft} & txt & 0.642 & -- & 7.4 & 10.85 \\
    DiT         & \textsc{ft} & vis & 0.623 & -- & 6.6 & 10.48 \\
    \midrule
    Phi-3-mini  & \textsc{zs} & txt & 0.060 & -- & 0.0 & 14.79 \\
    Mistral-7B  & \textsc{zs} & txt & 0.368 & -- & 1.4 & 21.15 \\
    GPT-4o$^{\mathrm{a}}$ & \textsc{zs} & txt & 0.703 & -- & 9.4 & 7.79 \\
    C-H4.5 (BD)$^{\mathrm{b}}$ & \textsc{zs} & txt & -- & 0.866 & -- & -- \\
    Gemini-3.1-FL (MSBD)$^{\mathrm{b}}$ & \textsc{zs} & vis & -- & 0.887 & -- & -- \\
    GPT-5.4-mini (BD)$^{\mathrm{b}}$ & \textsc{zs} & vis & -- & 0.876 & -- & -- \\
    \midrule
    Phi-3-mini  & \textsc{ft} & txt & 0.933 & -- & 63.7 & 1.76 \\
    Mistral-7B  & \textsc{ft} & txt & \textbf{0.967} & -- & \textbf{80.0} & \textbf{0.81} \\
    \midrule
    C-H4.5$^{\mathrm{250}}$ (RICoT) & \textsc{zs} & txt & 0.816 & 0.901 & 33.6 & 2.88 \\
    G5$^{\mathrm{250}}$ (RICoT) & \textsc{zs} & txt & \underline{0.847} & \textbf{0.915} & 41.6 & 2.40 \\
    Q3.5-2B$^{\star\,\mathrm{F}}$ (RICoT) & \textsc{dist} & txt & 0.774 & 0.890 & 23.2 & 4.21 \\
    Q3.5-2B$^{\star,\mathrm{PwT}\,\mathrm{F}}$ (RICoT) & \textsc{dist} & txt & 0.770 & 0.887 & 21.9 & 4.30 \\
    \bottomrule
  \end{tabular}
  \end{adjustbox}
\caption{TABME comparison against published baselines
  \citep{tabmepp,heidenreich-etal-2025-page,pottanigari2026msbd}.
  \textsc{zs}~/~\textsc{ft}: zero-shot vs.\ fine-tuned on TABME
  training split; \textsc{dist}: distilled from teacher outputs on a FinePDFs subset,
  no TABME data. Input: OCR text (txt) or page images (vis). Bold:
  overall best; underline: best zero-shot;
  --: not reported.
  $^{\mathrm{a}}$~GPT-4o-ZS evaluated on a $10\%$ subset;
  $^{\mathrm{b}}$~best result per model over all methods, window
  sizes, and inputs in~\citet{pottanigari2026msbd} (TABME++; BD:
  pairwise);
  $^{\mathrm{250}}$~TABME-250;
  $^{\mathrm{F}}$~full $5{,}000$-stream test set.}
  \label{tab:prior-tabme}
\end{table}

\subsection{RICoT vs.\ pairwise and windowed paradigms}
\label{sec:results-baselines}

Table~\ref{tab:baselines} reports DocF1 head-to-head against pairwise
and windowed LLM-PSS baselines on all three datasets for the GLM
family, with every method run on the same backbone (Tobacco800 has
only one stream, but the single-stream DocF1 still differentiates
configurations). On FinancePSS, RICoT is the best configuration for
every model, by $+0.154$, $+0.069$, and $+0.057$ over the best
baseline (GLM-4.7-flash, GLM-4.7, GLM-5). Without expand the margins
are $+0.133$ and $+0.032$ and a tie on GLM-5
(Section~\ref{sec:analysis}). Context length
does not explain the gain: in a control that keeps pairwise
decisions but shows a symmetric $5$-page context (GLM-4.7-flash,
$5$ runs; Appendix~\ref{app:matched-ctx}), pairwise rises from
$0.506$ to $0.651$, yet RICoT still leads it by $+0.111$ ($+0.090$
without expand) with $2.15$ vs.\ $\sim$$7.6$ calls per stream. On the near-ceiling Tobacco800, RICoT wins for
GLM-4.7 and GLM-5. On TABME-250, RICoT leads for GLM-4.7
($+0.052$) and GLM-5 ($+0.092$), while for the small GLM-4.7-flash,
pairwise classification with internal thinking is stronger ($0.771$
vs.\ $0.683$), mirroring its Tobacco800 result.

\begin{table}[t]
  \centering
  \scriptsize
  \setlength{\tabcolsep}{2.4pt}
  \begin{adjustbox}{max width=\columnwidth}
    \begin{tabular}{@{}lccc@{\hspace{5pt}}ccc@{\hspace{5pt}}ccc@{}}
    \toprule
    & \multicolumn{3}{c}{\textit{Tobacco800}} & \multicolumn{3}{c}{\textit{FinancePSS-205}} & \multicolumn{3}{c}{\textit{TABME-250}} \\
    \cmidrule(lr){2-4}\cmidrule(lr){5-7}\cmidrule(l){8-10}
    Method & G4.7-F & G4.7 & G5 & G4.7-F & G4.7 & G5 & G4.7-F & G4.7 & G5 \\
    \midrule
    Pairwise$^{\mathrm{iT}}$ & \textbf{0.886} & 0.888 & 0.872 & 0.506 & 0.763 & 0.739 & \textbf{0.771} & 0.704 & 0.697 \\
    Windowed$^{\mathrm{iT}}$ & 0.823 & 0.867 & 0.958 & 0.608 & 0.757 & 0.769 & 0.731 & 0.742 & 0.755 \\
    \textbf{RICoT} & 0.867 & \textbf{0.961} & \textbf{0.967} & \textbf{0.762} & \textbf{0.832} & \textbf{0.826} & 0.683 & \textbf{0.794} & \textbf{0.847} \\
    \quad no expand & -- & -- & -- & 0.741 & 0.795 & 0.769 & -- & -- & -- \\
    \bottomrule
  \end{tabular}
  \end{adjustbox}
\caption{DocF1 of RICoT vs.\ pairwise and windowed LLM-PSS
  baselines on the same backbone. Baselines run with internal
  thinking (their best general configuration); RICoT without
  (Section~\ref{sec:thinking-ablation}). Expand applies to
  FinancePSS only. Bold: best per column.}
  \label{tab:baselines}
\end{table}

\subsection{Pagewise vs.\ internal extended thinking}
\label{sec:thinking-ablation}

Do RICoT's gains overlap with models' internal extended thinking? We run a $2{\times}2$ ablation
on GLM-4.7-flash crossing pagewise interleaved thinking with
internal extended thinking across all three datasets
(Table~\ref{tab:thinking}). Pagewise yields a positive DocF1 delta
on every dataset; internal-thinking alone is positive only on
FinancePSS and neutral-to-negative elsewhere; the two are roughly
additive when combined. Per-page output-token costs differ by an
order of magnitude: on FinancePSS-flash, pagewise adds $\sim$$60$
output tokens to gain $+0.074$ DocF1, while internal thinking adds
$\sim$$650$ tokens for $+0.028$. The per-channel cost breakdown is
in Appendix~\ref{app:token-cost}. We therefore default to
internal-thinking-off for all RICoT runs in this paper.

\begin{table}[t]
  \centering
  \footnotesize
  \setlength{\tabcolsep}{4pt}
    \begin{adjustbox}{max width=\linewidth}
\begin{tabular}{@{}lcccc@{}}
    \toprule
    Configuration & FinPSS & Tob. & TABME & Out\,toks \\
    \midrule
    no-think (baseline) & 0.688 & 0.822 & 0.661 & 172 \\
    internal only       & $+0.028$ & $-0.012$ & $-0.008$ & 823 \\
    pagewise only       & $+0.074$ & $+0.045$ & $+0.022$ & 232 \\
    both                & $\mathbf{+0.118}$ & $\mathbf{+0.057}$ & $\mathbf{+0.046}$ & 722 \\
    \bottomrule
  \end{tabular}
  \end{adjustbox}
\caption{$2{\times}2$ DocF1 deltas vs.\ no-think baseline on
  GLM-4.7-flash. Last column reports per-page output tokens on
  FinancePSS-flash.}
  \label{tab:thinking}
\end{table}

\subsection{Component ablations}
\label{sec:results-ablations}

Table~\ref{tab:ablations} ablates each load-bearing component of
RICoT on FinancePSS-205 across the three prompted GLM models by
removing one component at a time and re-running. \emph{Pagewise
thinking} is the largest single contributor on all three models
($-0.048$ to $-0.088$ DocF1). \emph{Context propagation}
(reasoning history $\mathcal{H}^{(c)}$ and active-segment summary
$\mathcal{A}^{(c)}$) matters most for GLM-5 ($-0.048$) and on long
streams (Appendix~\ref{app:stratified}). \emph{Grounding} is
reliably positive for GLM-4.7-flash and GLM-5. Removing only the
model-emitted \emph{carry-forward block} is mixed, so the
state-propagation story rests on the reasoning history;
Appendix~\ref{app:uncertainty} marks which deltas exceed run-to-run
noise. \emph{Expand} contributes
$+0.021$ to $+0.057$. Per-dataset full
ablation tables, including TABME and Tobacco800, are in
Appendices~\ref{app:abl-thinking}, \ref{app:abl-state},
and~\ref{app:abl-decision}.

\begin{table}[t]
  \centering
  \scriptsize
  \setlength{\tabcolsep}{3pt}
    \begin{tabular}{@{}lccc@{}}
    \toprule
    Configuration & G4.7-F & G4.7 & G5 \\
    \midrule
    RICoT (full)            & 0.762 & 0.832 & 0.826 \\
    \midrule
    $-$ pagewise thinking   & 0.688\,($-.074$) & 0.784\,($-.048$) & 0.738\,($-.088$) \\
    $-$ context propagation & 0.752\,($-.010$) & 0.837\,($+.005$) & 0.778\,($-.048$) \\
    $-$ carry-forward block & 0.736\,($-.026$) & 0.841\,($+.009$) & 0.795\,($-.031$) \\
    $-$ grounding           & 0.695\,($-.067$) & 0.831\,($-.001$) & 0.755\,($-.071$) \\
    $-$ expand (no-expand)  & 0.741\,($-.021$) & 0.795\,($-.037$) & 0.769\,($-.057$) \\
    \bottomrule
  \end{tabular}
\caption{Component ablations on FinancePSS-205 DocF1. Each row
  removes one component from the full RICoT (with-expand)
  configuration; $\Delta$ shown in parentheses. ``Carry-forward
  block'' removes only the model-emitted $\kappa$.}
  \label{tab:ablations}
\end{table}

\subsection{Discussion}
\label{sec:analysis}

\paragraph{Segment granularity on TABME.}
TABME assigns one segment to each source file, whereas RICoT
segments individual document instances and marks nested documents
on its sub-document axis. Consistent with this difference, RICoT's
TABME errors are predominantly additional boundaries: page-level
recall is at least $0.955$ for every prompted model, against
precision of $0.733$--$0.873$ (Appendix~\ref{app:tabme-pr}). To
characterise these additional boundaries, we manually reviewed
$100$ false boundaries of Q3.5-2B$^{\star}$ on the full test set
(Appendix~\ref{app:tabme-errors}). Weighted by how often each error
recurs across streams, about $17\%$ are genuine document starts
within a single source file, which file-level labels merge
(treating them as correct would raise precision from $0.836$ to
about $0.86$); about $33\%$ are nested documents, such as envelopes,
attachments, and fax payloads, predicted as top-level rather than
sub-document starts; about $33\%$ are over-segmentations triggered
by repeated headers or explicit continuations (``Page~2 of~2''); and
the remaining $17\%$ are ambiguous. The precision gap on TABME thus
reflects both a difference in segment granularity and genuine model
errors, and fine-tuning on TABME's training split plausibly
addresses both.

\paragraph{Expand on FinancePSS.}
On FinancePSS, expand drives a meaningful share of the gain for
the larger models: without it, RICoT-GLM-5 ties the best baseline
and RICoT-GLM-4.7 leads by $+0.032$; with it the margins become
$+0.057$ and $+0.069$. GLM-4.7-flash wins regardless ($+0.133$
without expand). GLM-5 more often places bundled-packet boundaries
on the sub-document axis rather than the segment axis; expand
re-thresholds these detected starts, closing its gap to GLM-4.7
from $0.026$ to $0.006$.

\section*{Limitations}

Our evaluation is restricted to English documents; multilingual streams---common in global enterprises---remain untested, and we did not evaluate on recent non-English open benchmarks~\cite{wooir,openpss}. The discriminating evidence rests mainly on FinancePSS-205, which is modest in size ($205$ streams), curated rather than randomly sampled, labeled by the research team without an agreement study, and covers a single domain, so generalization to legal or medical content requires further validation; Tobacco800 is near ceiling, and TABME scores file-level segments, so its precision mixes granularity differences with model errors (Section~\ref{sec:analysis}). Run-to-run variability was measured with five repeats on GLM-4.7-flash (Appendix~\ref{app:uncertainty}) and assumed to transfer to the other models, whose results are single runs; the student was trained once, and the effect of omitting optional decision fields during training was not isolated. Model coverage spans one model family at three scales plus one non-GLM model. The framework assumes that the first page of a stream begins a document and treats each page image as one logical unit, with no granularity for multiple pages combined into a single scan. It relies on sequential continuity and cannot reassociate non-contiguous or interleaved pages into one document entity. Finally, recurrent state propagation makes within-stream processing strictly causal: multiple streams can be processed in parallel, but a single long stream must be processed sequentially.

\section*{Ethical Considerations}

The FinancePSS dataset was collected from production workflows under the data-handling agreements in force at the time and pseudonymised: personal names, addresses, contact details, and account numbers were consistently replaced with type- and length-consistent synthetic values, preserving document structure. The planned release will contain only this de-identified text (no page images), which was audited for retained identifiers (Appendix~\ref{app:dataset-card}), and will be licensed for non-commercial research use (CC BY-NC 4.0) under terms of use that prohibit re-identification; because free-text pseudonymisation cannot fully exclude re-identification, we treat the release as pseudonymised rather than anonymous. No human annotators beyond the research team were employed for dataset labeling. AI writing assistance was used for portions of manuscript drafting; all scientific content, experimental design, results, and conclusions are entirely the authors' own.

\bibliography{custom}

\begin{thebibliography}{31}
\providecommand{\natexlab}[1]{#1}

\bibitem[{Agam et~al.(2006)Agam, Argamon, Frieder, Grossman, and
  Lewis}]{agam2006cdip}
Gady Agam, Shlomo Argamon, Ophir Frieder, David Grossman, and David~D. Lewis.
  2006.
\newblock \href {http://ir.iit.edu/projects/CDIP.html} {The complex document
  image processing ({CDIP}) test collection}.
\newblock Technical report, Illinois Institute of Technology.

\bibitem[{{Anthropic}(2025)}]{claude_haiku45}
{Anthropic}. 2025.
\newblock Introducing {Claude Haiku 4.5}.
\newblock Anthropic blog post.
  \url{https://www.anthropic.com/news/claude-haiku-4-5}.

\bibitem[{Beltagy et~al.(2020)Beltagy, Peters, and
  Cohan}]{beltagy2020longformer}
Iz~Beltagy, Matthew~E. Peters, and Arman Cohan. 2020.
\newblock Longformer: The long-document transformer.
\newblock \emph{arXiv preprint arXiv:2004.05150}.

\bibitem[{Braz et~al.(2021)Braz, da~Silva, and Lima}]{braz2021leveraging}
Fabricio~Ataides Braz, Nilton~Correia da~Silva, and Jonathan Alis~Salgado Lima.
  2021.
\newblock \href {https://doi.org/10.1016/j.engappai.2021.104394} {Leveraging
  effectiveness and efficiency in {Page Stream Deep Segmentation}}.
\newblock \emph{Engineering Applications of Artificial Intelligence},
  105:104394.

\bibitem[{Dettmers et~al.(2023)Dettmers, Pagnoni, Holtzman, and
  Zettlemoyer}]{qlora}
Tim Dettmers, Artidoro Pagnoni, Ari Holtzman, and Luke Zettlemoyer. 2023.
\newblock \href {https://arxiv.org/abs/2305.14314} {{QLoRA}: Efficient
  finetuning of quantized {LLMs}}.
\newblock In \emph{Advances in Neural Information Processing Systems},
  volume~36.

\bibitem[{Drumond et~al.(2023)Drumond, Leite, de~Campos, and
  Braz}]{drumond2023layoutqt}
Patricia Medyna Lauritzen de~Lucena Drumond, Lindeberg~Pessoa Leite, Teofilo~E.
  de~Campos, and Fabricio~Ataides Braz. 2023.
\newblock \href {https://doi.org/10.1016/j.engappai.2023.106091}
  {{LayoutQT}---{Layout Quadrant Tags} to embed visual features for document
  analysis}.
\newblock \emph{Engineering Applications of Artificial Intelligence},
  122:106091.

\bibitem[{{GLM-5 Team}(2026)}]{glm5}
{GLM-5 Team}. 2026.
\newblock \href {https://arxiv.org/abs/2602.15763} {{GLM-5}: from vibe coding
  to agentic engineering}.
\newblock \emph{arXiv preprint arXiv:2602.15763}.

\bibitem[{{GLM Team} et~al.(2025){GLM Team}, Zeng et~al.}]{glm}
{GLM Team}, Aohan Zeng, et~al. 2025.
\newblock \href {https://arxiv.org/abs/2508.06471} {{GLM-4.5}: Agentic,
  reasoning, and coding ({ARC}) foundation models}.
\newblock \emph{arXiv preprint arXiv:2508.06471}.
\newblock Z.ai. The GLM-4.7 family used in our experiments is a subsequent
  release in the same line; see \url{https://github.com/zai-org/GLM-4.5} for
  model cards.

\bibitem[{Heidenreich et~al.(2024)Heidenreich, Dalvi, Mukku, Verma, and
  Pi{\v{c}}uljan}]{tabmepp}
Hunter Heidenreich, Ratish Dalvi, Rohith Mukku, Nikhil Verma, and Neven
  Pi{\v{c}}uljan. 2024.
\newblock \href {https://arxiv.org/abs/2408.11981} {Large language models for
  page stream segmentation}.
\newblock \emph{Preprint}, arXiv:2408.11981.

\bibitem[{Heidenreich et~al.(2025)Heidenreich, Dalvi, Verma, and
  Getachew}]{heidenreich-etal-2025-page}
Hunter Heidenreich, Ratish Dalvi, Nikhil Verma, and Yosheb Getachew. 2025.
\newblock \href {https://aclanthology.org/2025.coling-industry.26/} {Page
  stream segmentation with {LLM}s: Challenges and applications in insurance
  document automation}.
\newblock In \emph{Proceedings of the 31st International Conference on
  Computational Linguistics: Industry Track}, pages 305--317, Abu Dhabi, UAE.
  Association for Computational Linguistics.

\bibitem[{Inan et~al.(2022)Inan, Rungta, and Mehdad}]{inan2022structured}
Hakan Inan, Rashi Rungta, and Yashar Mehdad. 2022.
\newblock \href {https://arxiv.org/abs/2209.13759} {Structured summarization:
  Unified text segmentation and segment labeling as a generation task}.
\newblock \emph{Preprint}, arXiv:2209.13759.

\bibitem[{{Kimi Team}(2025)}]{kimi_k2}
{Kimi Team}. 2025.
\newblock \href {https://arxiv.org/abs/2507.20534} {{Kimi K2}: Open agentic
  intelligence}.
\newblock \emph{arXiv preprint arXiv:2507.20534}.
\newblock Base and post-trained checkpoints at
  \url{https://huggingface.co/moonshotai/Kimi-K2-Instruct}.

\bibitem[{Kydl{\'\i}{\v{c}}ek et~al.(2025)Kydl{\'\i}{\v{c}}ek, Penedo, and von
  Werra}]{kydlicek2025finepdfs}
Hynek Kydl{\'\i}{\v{c}}ek, Guilherme Penedo, and Leandro von Werra. 2025.
\newblock {FinePDFs}.
\newblock Hugging Face dataset,
  \url{https://huggingface.co/datasets/HuggingFaceFW/finepdfs}.

\bibitem[{Lewis et~al.(2006)Lewis, Agam, Argamon, Frieder, Grossman, and
  Heard}]{tobacco800}
David~D. Lewis, Gady Agam, Shlomo Argamon, Ophir Frieder, David Grossman, and
  John Heard. 2006.
\newblock Building a test collection for complex document information
  processing.
\newblock In \emph{Proceedings of the 29th Annual International ACM SIGIR
  Conference on Research and Development in Information Retrieval (SIGIR '06)},
  pages 665--666. ACM.

\bibitem[{Li et~al.(2026)Li, Shen, Jeoung, Chen, Li, Zhu, Wang, Ioannidis, and
  Rangwala}]{li2026boundrl}
Haoyuan Li, Zhengyuan Shen, Sullam Jeoung, Yueyan Chen, Jiayu Li, Qi~Zhu, Shuai
  Wang, Vassilis Ioannidis, and Huzefa Rangwala. 2026.
\newblock {BoundRL}: Efficient token-level structured text segmentation through
  reinforced boundary generation.
\newblock In \emph{Findings of the Association for Computational Linguistics:
  ACL 2026}, pages 34706--34726. Association for Computational Linguistics.

\bibitem[{Liu et~al.(2024)Liu, Lin, Hewitt, Paranjape, Bevilacqua, Petroni, and
  Liang}]{liu2024lost}
Nelson~F. Liu, Kevin Lin, John Hewitt, Ashwin Paranjape, Michele Bevilacqua,
  Fabio Petroni, and Percy Liang. 2024.
\newblock Lost in the middle: How language models use long contexts.
\newblock \emph{Transactions of the Association for Computational Linguistics},
  12:157--173.

\bibitem[{Mungmeeprued et~al.(2022)Mungmeeprued, Ma, Mehta, and
  Lipani}]{tabme2022}
Thisanaporn Mungmeeprued, Yuxin Ma, Nisarg Mehta, and Aldo Lipani. 2022.
\newblock \href {https://doi.org/10.1145/3558100.3563852} {Tab this folder of
  documents: page stream segmentation of business documents}.
\newblock In \emph{Proceedings of the 22nd ACM Symposium on Document
  Engineering}, DocEng '22, New York, NY, USA. Association for Computing
  Machinery.

\bibitem[{Packer et~al.(2023)Packer, Wooders, Lin, Fang, Patil, Stoica, and
  Gonzalez}]{packer2023memgpt}
Charles Packer, Sarah Wooders, Kevin Lin, Vivian Fang, Shishir~G. Patil, Ion
  Stoica, and Joseph~E. Gonzalez. 2023.
\newblock {MemGPT}: Towards {LLMs} as operating systems.
\newblock \emph{arXiv preprint arXiv:2310.08560}.

\bibitem[{Pottanigari et~al.(2026)Pottanigari, Kharaghani, Vadacchino, Posada,
  and Zhang}]{pottanigari2026msbd}
Nikhil~Reddy Pottanigari, Sepideh Kharaghani, Saverio Vadacchino, Alejandro
  Posada, and Ying Zhang. 2026.
\newblock \href {https://arxiv.org/abs/2609.22620} {Splitting documents at
  lower cost: Multi-split boundary decisions for {LLM}-based page stream
  segmentation}.
\newblock \emph{Preprint}, arXiv:2609.22620.

\bibitem[{{Qwen Team}(2026)}]{qwen}
{Qwen Team}. 2026.
\newblock {Qwen3.5}: Model release.
\newblock \url{https://huggingface.co/Qwen/Qwen3.5-2B}.
\newblock Model card for Qwen3.5-2B.

\bibitem[{Serra~Ortega et~al.(2025)Serra~Ortega, Vivoli, Llabr\'{e}s, and
  Karatzas}]{serraortega2025cosmo}
Marc Serra~Ortega, Emanuele Vivoli, Artemis Llabr\'{e}s, and Dimosthenis
  Karatzas. 2025.
\newblock \href {https://arxiv.org/abs/2507.10053} {{CoSMo}: A multimodal
  transformer for page stream segmentation in comic books}.
\newblock In \emph{Proceedings of the IEEE/CVF International Conference on
  Computer Vision (ICCV) Workshops}, pages 7506--7514.

\bibitem[{{University of California, San Francisco}(2007)}]{ltdl2007}
{University of California, San Francisco}. 2007.
\newblock The legacy tobacco document library ({LTDL}).
\newblock Now the UCSF Industry Documents Library,
  \url{https://www.industrydocuments.ucsf.edu/}.

\bibitem[{van Heusden et~al.(2022)van Heusden, Kamps, and Marx}]{wooir}
Ruben van Heusden, Jaap Kamps, and Maarten Marx. 2022.
\newblock {WooIR}: A new open page stream segmentation dataset.
\newblock In \emph{Proceedings of the 2022 ACM SIGIR International Conference
  on the Theory of Information Retrieval (ICTIR)}.

\bibitem[{van Heusden et~al.(2024)van Heusden, Kamps, and Marx}]{openpss}
Ruben van Heusden, Jaap Kamps, and Maarten Marx. 2024.
\newblock \href {https://doi.org/10.1007/978-3-031-72437-4_24} {{OpenPSS}: An
  open page stream segmentation benchmark}.
\newblock In \emph{Linking Theory and Practice of Digital Libraries (TPDL
  2024)}, volume 15177 of \emph{Lecture Notes in Computer Science}, pages
  413--429. Springer.

\bibitem[{Wei et~al.(2022)Wei, Wang, Schuurmans, Bosma, Ichter, Xia, Chi, Le,
  and Zhou}]{wei2022chain}
Jason Wei, Xuezhi Wang, Dale Schuurmans, Maarten Bosma, Brian Ichter, Fei Xia,
  Ed~Chi, Quoc Le, and Denny Zhou. 2022.
\newblock Chain-of-thought prompting elicits reasoning in large language
  models.
\newblock In \emph{Advances in Neural Information Processing Systems},
  volume~35, pages 24824--24837.

\bibitem[{Wiedemann and Heyer(2018)}]{wiedemann2018}
Gregor Wiedemann and Gerhard Heyer. 2018.
\newblock Multi-modal page stream segmentation with convolutional neural
  networks.
\newblock In \emph{Proceedings of the 27th ACM International Conference on
  Information and Knowledge Management}, pages 1735--1738. ACM.

\bibitem[{Wiedemann and Heyer(2020)}]{wiedemann_heyer}
Gregor Wiedemann and Gerhard Heyer. 2020.
\newblock \href {https://doi.org/10.1007/s10579-019-09476-2} {Multi-modal page
  stream segmentation with convolutional neural networks}.
\newblock \emph{Language Resources and Evaluation}, 54(4):867--888.

\bibitem[{Xie et~al.(2025)Xie, Qiu, Gopinath, Lin, Sun, Wang, Potdar, and
  Dhingra}]{xie2026interleaved}
Roy Xie, David Qiu, Deepak Gopinath, Dong Lin, Yanchao Sun, Chong Wang, Saloni
  Potdar, and Bhuwan Dhingra. 2025.
\newblock \href {https://arxiv.org/abs/2505.19640} {Interleaved reasoning for
  large language models via reinforcement learning}.
\newblock \emph{Preprint}, arXiv:2505.19640.

\bibitem[{Yao et~al.(2023)Yao, Zhao, Yu, Du, Shafran, Narasimhan, and
  Cao}]{yao2023react}
Shunyu Yao, Jeffrey Zhao, Dian Yu, Nan Du, Izhak Shafran, Karthik Narasimhan,
  and Yuan Cao. 2023.
\newblock {ReAct}: Synergizing reasoning and acting in language models.
\newblock In \emph{International Conference on Learning Representations}.

\bibitem[{Zaheer et~al.(2020)Zaheer, Guruganesh, Dubey, Ainslie, Alberti,
  Ontanon, Pham, Ravula, Wang, Yang et~al.}]{zaheer2020bigbird}
Manzil Zaheer, Guru Guruganesh, Kumar~Avinava Dubey, Joshua Ainslie, Chris
  Alberti, Santiago Ontanon, Philip Pham, Anirudh Ravula, Qifan Wang, Li~Yang,
  et~al. 2020.
\newblock Big bird: Transformers for longer sequences.
\newblock In \emph{Advances in Neural Information Processing Systems},
  volume~33, pages 17283--17297.

\bibitem[{Zhang et~al.(2025)Zhang, Kraska, and Khattab}]{zhang2025recursive}
Alex~L. Zhang, Tim Kraska, and Omar Khattab. 2025.
\newblock \href {https://arxiv.org/abs/2512.24601} {Recursive language models}.
\newblock \emph{Preprint}, arXiv:2512.24601.

\end{thebibliography}

\appendix
\raggedbottom 

\section{FinancePSS-205 Dataset Card}
\label{app:dataset-card}

\paragraph{Access and license.} The dataset is not yet public; it
will be released together with the evaluation code, and the URL
will be added in a future version of this preprint. Text and
annotations will be licensed under CC BY-NC 4.0 for non-commercial
research use, with separate terms of use prohibiting any attempt at
re-identification; evaluation scripts under Apache-2.0.

\paragraph{Source and composition.} $205$ page streams ($1{,}772$
pages, $685$ segments) as they arrived in production
accounts-payable/receivable workflows; documents date from roughly
2024--2025. Each stream is one real bundle: there is no
concatenation or reuse, and no stream is duplicated. The set is a
curated selection, favouring streams that were fully processable
and whose de-identification could be verified, so it is not an
unbiased sample of the production distribution. Segment labels are
free text with $144$ distinct document types
(Table~\ref{tab:fin-classes}).

\paragraph{Format.} One UTF-8 text file per page in stream order,
containing layout-preserving OCR text (no images, coordinates, or
PDFs); \texttt{financepss\_gt.json} lists each stream's segments as
(\texttt{start\_page}, \texttt{page\_count},
\texttt{document\_label}), contiguous and tiling the stream;
\texttt{MANIFEST.json} gives per-file SHA-256 prefixes. OCR noise is
not corrected; $30$ pages ($1.7\%$) are empty after OCR.

\paragraph{Annotation.} A new segment starts at the first page of
each functionally distinct document. Labels were produced by the
research team; no inter-annotator agreement is available for this
version. The release encodes only the segment axis; sub-document
annotations are planned for a later version.

\paragraph{De-identification.} Personal names, addresses, contact
details, and account and instrument numbers were replaced with
type- and length-consistent surrogates, so identifier changes, a
primary boundary cue, survive. An audit on $25$ sampled streams
found mean character similarity to the source of $0.579$ with file
lengths within $0.5\%$; Luhn-valid $13$--$19$-digit strings occur at
the chance rate ($10.2\%$), indicating no retained card numbers. The
release is pseudonymised rather than anonymous, and residual
re-identification risk is not zero.

\paragraph{Difficulty.} $48.3\%$ of streams contain a single
document, but these hold few of the documents: $46.7\%$ of segments
are single pages and the longest stream has $44$. Trivial
predictors therefore score low (Table~\ref{tab:trivial}).

\begin{table}[H]
  \centering
  \small
  \begin{adjustbox}{max width=\columnwidth}
  \begin{tabular}{lcc}
    \toprule
    Predictor & DocF1 & B-F1 \\
    \midrule
    Every page a document & 0.260 & 0.469 \\
    Whole stream one document & 0.222 & 0.000 \\
    RICoT, G4.7 (no expand) & 0.795 & -- \\
    \bottomrule
  \end{tabular}
  \end{adjustbox}
\caption{Trivial predictors versus RICoT on FinancePSS-205
  (micro-averaged).}
  \label{tab:trivial}
\end{table}

\begin{table}[H]
  \centering
  \small
  \begin{adjustbox}{max width=\columnwidth}
  \begin{tabular}{lr}
    \toprule
    Document class & Segments \\
    \midrule
    invoice & 330 \\
    receipt & 35 \\
    email & 18 \\
    bill of lading & 18 \\
    statement & 15 \\
    purchase order & 15 \\
    delivery receipt & 15 \\
    packing list & 12 \\
    commercial invoice & 9 \\
    hazmat driver checklist & 9 \\
    other ($134$ types) & 209 \\
    \bottomrule
  \end{tabular}
  \end{adjustbox}
\caption{FinancePSS-205 document-class distribution
  (case-normalised labels).}
  \label{tab:fin-classes}
\end{table}

\section{Uncertainty Estimates}
\label{app:uncertainty}

\paragraph{Run-to-run variability.} Prompted results in the main
paper are single runs, since multi-run coverage of every
model$\times$configuration was cost-prohibitive. To measure residual
provider-side variability, we repeated the GLM-4.7-flash RICoT
configuration $4$ more times with identical code, configuration,
and decoding (temperature $0.001$, top-$p$ $1$).
Table~\ref{tab:ci-runs} reports the $5$-run means (paper run plus
$4$ repeats) with $95\%$ CIs (Student's $t$, $n{=}5$). The runs are
genuinely independent: $92$ of $205$ FinancePSS streams receive
different split counts across repeats. Every value reported in the
main paper lies inside its CI. We treat differences smaller than
the DocF1 band as inconclusive.

\begin{table}[h]
  \centering
  \scriptsize
  \setlength{\tabcolsep}{2.5pt}
  \begin{adjustbox}{max width=\columnwidth}
  \begin{tabular}{@{}lccccc@{}}
    \toprule
    Dataset & DocF1 & PageF1 & $\mathrm{MNDD}_\mu$ & STP\% & Paper \\
    \midrule
    FinPSS (exp.)    & $0.751{\pm}0.026$ & $0.890{\pm}0.017$ & $1.16{\pm}0.24$ & $67.1{\pm}1.6$ & 0.762 \\
    FinPSS (no exp.) & $0.739{\pm}0.012$ & $0.878{\pm}0.013$ & $1.32{\pm}0.19$ & $65.7{\pm}1.5$ & 0.741 \\
    Tobacco800       & $0.844{\pm}0.053$ & $0.928{\pm}0.036$ & $43.0{\pm}14.6$ & $0.0$ & 0.867 \\
    TABME-250        & $0.686{\pm}0.006$ & $0.833{\pm}0.007$ & $5.65{\pm}0.26$ & $16.2{\pm}2.3$ & 0.683 \\
    \bottomrule
  \end{tabular}
  \end{adjustbox}
\caption{GLM-4.7-flash RICoT, $5$-run mean $\pm$ $95\%$ CI. ``Paper'':
  the single-run DocF1 reported in the main tables. Tobacco800 is one
  $259$-page stream, hence its wider band.}
  \label{tab:ci-runs}
\end{table}

\paragraph{Implications for the main results.}
\begin{itemize}
\setlength{\itemsep}{0pt}
\item \emph{Table~\ref{tab:baselines}, Tobacco800:} only the GLM-4.7
margin over the best baseline ($+0.073$) exceeds the $\pm0.053$
band; Tobacco800 margins are otherwise inconclusive.
\item \emph{Matched-context pairwise control
(Appendix~\ref{app:matched-ctx}):}
$0.651\pm0.008$ over $5$ runs; RICoT's lead ($+0.111$ with expand,
$+0.090$ without) is far outside both this spread and the
run-to-run band.
\item \emph{Table~\ref{tab:ablations}:} deltas with
$|\Delta|\le0.026$ are within noise: context propagation on
GLM-4.7-flash ($-0.010$) and GLM-4.7 ($+0.005$), carry-forward
block on GLM-4.7-flash ($-0.026$) and GLM-4.7 ($+0.009$), grounding
on GLM-4.7 ($-0.001$), and expand on GLM-4.7-flash ($-0.021$). The
band was measured on GLM-4.7-flash and is assumed to transfer.
\end{itemize}

\paragraph{Bootstrap over streams.} $95\%$ CIs ($10$k resamples of
streams) for the strongest configuration on each multi-stream
dataset: FinancePSS with expand, GLM-4.7 $0.832$ [$0.782$--$0.878$]
and GLM-5 $0.826$ [$0.773$--$0.871$]; TABME-250, GLM-5 $0.847$
[$0.826$--$0.868$]. Tobacco800 is a single stream, so its
uncertainty is captured by the run-to-run band above.

\paragraph{Source-file reuse on TABME.} TABME composes its
$5{,}000$ test streams from $2{,}238$ source files, each used in
$24.5$ streams on average (range $9$--$42$); this reuse is what lets
a fixed archive yield a large and varied test set. A model's error
inside a file therefore recurs wherever the file appears: for
Q3.5-2B$^{\star}$ on the full test set, the $9{,}331$ false
boundaries fall on only $1{,}071$ distinct (file, page) sites, and
the $10\%$ of files with the most false boundaries hold $87\%$ of
them. Streams are thus not independent units. Resampling streams
($10$k resamples) gives B-F1 $0.890$ [$0.887$--$0.893$], while
resampling source files gives [$0.877$--$0.902$], about four times
wider; counting each source file once gives B-F1 $0.896$. We
therefore recommend file-level resampling for model comparisons on
the full TABME test set. On TABME-250 reuse is much lower
($2{,}627$ segments from $1{,}542$ unique files;
Table~\ref{tab:datasets}), so stream-level intervals there are less
affected.

\paragraph{Distilled student ($5$ inference runs).} The student was
trained once. FinancePSS DocF1 with expand: $0.773$
($0.763$--$0.782$) for Q3.5-2B$^{\star}$ and $0.785$
($0.772$--$0.797$) for Q3.5-2B$^{\star,\mathrm{PwT}}$;
Tobacco800: $0.914$ ($0.897$--$0.930$) and $0.927$
($0.898$--$0.956$).

\section{Window-Parameter Sweep}
\label{app:window-sweep}

On GLM-4.7-flash / FinancePSS we replaced RICoT's token-budget
packing with static windows of $w$ pages and overlap $o$, one run
per configuration (Table~\ref{tab:window-sweep}). \emph{Overlap $1$
is best:} at $w{=}4$, raising $o$ from $1$ to $3$ increases cost
from $3.16$ to $8.50$ calls per stream while DocF1 falls from
$0.763$ to $0.728$. \emph{Smaller windows are more accurate but
costlier:} $w{=}2$ reaches $0.806$ ($0.811$ without expand) at about
$4\times$ the calls of the default, which trades $0.03$--$0.06$
DocF1 for that cost; it was chosen for cost, not tuned for accuracy.
Since each page is capped at $800$ tokens, $B{=}3600$ admits at
least $4$ pages, and $w_{\max}{=}6$ bounds the window from above, so
the effective window floats within this band. The token-budget
setting reaches $0.777$ at $2.15$ calls per stream, matching static
$w{=}5$ ($0.780$) and beating static
$w{=}6$ ($0.766$) at identical cost. $\mathrm{MNDD}_\mu$ ranges only
$0.86$--$1.21$ across configurations, within its $\pm0.24$ noise,
so it does not discriminate between them.

\begin{table}[h]
  \centering
  \small
  \setlength{\tabcolsep}{4pt}
  \begin{adjustbox}{max width=\columnwidth}
  \begin{tabular}{@{}cccccc@{}}
    \toprule
    $w$ & $o$ & DocF1 & (no exp.) & PageF1 & calls \\
    \midrule
    2 & 1 & \textbf{0.806} & 0.811 & 0.917 & 8.50 \\
    3 & 1 & 0.788 & 0.768 & 0.909 & 4.44 \\
    4 & 1 & 0.763 & 0.760 & 0.897 & 3.16 \\
    4 & 2 & 0.762 & 0.766 & 0.892 & 4.44 \\
    4 & 3 & 0.728 & 0.715 & 0.891 & 8.50 \\
    5 & 1 & 0.780 & 0.750 & 0.904 & 2.49 \\
    5 & 2 & 0.770 & 0.751 & 0.900 & 3.16 \\
    5 & 3 & 0.777 & 0.739 & 0.904 & 4.44 \\
    6 & 1 & 0.766 & 0.749 & 0.887 & 2.15 \\
    7 & 1 & 0.760 & 0.745 & 0.897 & 1.92 \\
    \midrule
    \multicolumn{2}{c}{$B{=}3600$, $w_{\max}{=}6$} & 0.777 & 0.755 & 0.907 & 2.15 \\
    \bottomrule
  \end{tabular}
  \end{adjustbox}
\caption{Window-parameter sweep, GLM-4.7-flash, FinancePSS-205,
  single runs. DocF1/PageF1 with expand; no-expand DocF1 in the
  fourth column. ``calls'': LLM calls per stream.}
  \label{tab:window-sweep}
\end{table}

\section{Matched-Context Pairwise Control}
\label{app:matched-ctx}

To test whether RICoT's gain is a context-length effect, we keep
the pairwise decision paradigm (one independent call per page,
predicting only that page's label, no carried state or joint
decoding) but show a symmetric $5$-page window $(n{-}2,\dots,n{+}2)$
instead of $(n{-}1,n)$. Pages use the same head-$500$/tail-$300$
truncation as the Windowed baseline, so per-decision context is
comparable to Windowed and RICoT windows. GLM-4.7-flash, internal
thinking on (as for the other baselines), temperature $0.001$;
$5$ full runs on FinancePSS-205, all $205$ streams with no parse
failures.

\paragraph{Results.} DocF1 $0.651\pm0.008$ (PageF1 $0.852\pm0.006$,
$95\%$ CI). Context matters: it lifts pairwise from $0.506$ to
$0.651$, which accounts for Windowed's advantage over pairwise
($0.608$). With context matched, RICoT still leads by $+0.111$
($+0.090$ without expand), far outside the CI.

\paragraph{Cost.} The control makes $7.6$ calls per stream (one per
page after the first, each re-sending up to $5$ pages) versus RICoT's $2.15$; each
call uses $2{,}860$ input and $1{,}277$ internal-thinking tokens on
average, whereas RICoT runs with internal thinking off.

\section{Long Streams and Error Propagation}
\label{app:stratified}

\paragraph{Depth-stratified state propagation.} On the $17$ longest
FinancePSS streams, restricting to deep pages (windows $\geq5$), full
RICoT outperforms the no-context-propagation ablation by $+0.178$
DocF1 (GLM-5) and $+0.131$ (GLM-4.7), while the effect is negative
for GLM-4.7-flash. Recurrent state therefore matters most
where it is exercised (long streams) and for models with the
capacity to use it; on short streams, which fit in one or two
windows, it has little to act on.

\paragraph{Error cascading.} Mapping boundary errors to windows
shows local persistence rather than systematic corruption: the
next-window error rate is $50$--$71\%$ after an erroneous window
versus $5$--$12\%$ after a correct one, yet $54$--$69\%$ of streams
with an early error later produce an error-free window (median
error run: $2$ windows). This analysis is observational; a causal
estimate would require replay with corrected state.

\section{Worked Inference Example}
\label{app:worked-example}

Below is the teacher output for the first window (pages 1--2 of 4
shown) of a FinePDFs stream, in the full schema used for
distillation. Per-page reasoning precedes each decision; the
grounding fingerprint names the identifiers that define the active
segment, and later pages cite fulfillment evidence against it.
Because this is the first window, no reasoning history or
active-segment summary is passed in; from the second window on,
$\mathcal{H}^{(c)}$ carries the most recent reasoning--decision
pairs in this same format, and $\mathcal{A}^{(c)}$ carries the
active segment's start page, class, pages covered, and the
model-emitted carry-forward block. The complete chunk, including
its two decisions-only renderings (Appendix~\ref{app:samples}), will be included in the public release, with
the verbatim instruction prose withheld (Appendix~\ref{app:prompts}).

\begin{tcolorbox}[breakable,boxsep=1pt,left=2pt,right=2pt,top=2pt,bottom=2pt,colback=white]
\ttfamily\scriptsize\raggedright
<analyze\_page> First page of the stream. Header clearly identifies
this as a SEC Form 10-K annual report for Landec Corporation, fiscal
year ended October 28, 2001, with commission file number 0-27446.
Active segment resets to SEC Form 10-K annual report grounded by
Landec Corporation, FY 2001, file number 0-27446. </analyze\_page>\\[2pt]
<decision> <page>1</page>\\
\quad<segment\_boundary> <type>SEGMENT\_START</type>\\
\quad<document\_class>SEC Form 10-K annual report</document\_class>\\
\quad<grounding> <fingerprint>Landec Corporation, FY ended
10/28/2001, Commission file number 0-27446</fingerprint>\\
\quad<initiation\_evidence>Header Dominance: SEC Form 10-K header
with clear company identification, fiscal year, and commission file
number; first page of stream</initiation\_evidence></grounding>\\
\quad<confidence>high</confidence></segment\_boundary>\\
\quad<page\_role>cover page with company information and securities
registration details</page\_role> </decision>\\[4pt]
<analyze\_page> Page 2 continues the same Form 10-K document with
``LANDEC CORPORATION ANNUAL REPORT ON FORM 10-K'' header and table
of contents listing Items 1-14. [\dots] No new segment initiation
signals present. </analyze\_page>\\[2pt]
<decision> <page>2</page>\\
\quad<segment\_boundary> <type>SEGMENT\_CONTINUATION</type>\\
\quad<grounding> <fulfillment\_evidence>Same Form 10-K document
continuation, table of contents follows cover page structure,
consistent company name and document type</fulfillment\_evidence>
</grounding>\\
\quad<confidence>high</confidence></segment\_boundary>\\
\quad<sub\_document\_start>none</sub\_document\_start>\\
\quad<page\_role>table of contents and beginning of Part
I</page\_role> </decision>
\end{tcolorbox}

\section{TABME-250 Stream IDs}
\label{app:tabme-streams}

The $250$ TABME stream IDs used for all prompting experiments were
sampled uniformly at random from TABME's $5{,}000$-stream test
split with a fixed seed and held constant across runs:

\begin{quote}\footnotesize\sloppy
2492, 1410, 1806, 1699, 2082, 2914, 1944, 3341, 4049, 4073, 2664,
2227, 1672, 2658, 2375, 2024, 2099, 758, 1374, 2002, 3178, 4357,
3601, 1295, 4428, 134, 3494, 2527, 4925, 176, 1965, 4676, 701,
1492, 2347, 4195, 2997, 4865, 225, 1994, 4981, 3060, 2037, 575,
3342, 1517, 3121, 705, 2449, 204, 3614, 440, 2233, 4297, 2162,
4230, 1863, 3107, 2826, 3273, 1089, 3823, 4807, 409, 2605, 2794,
2499, 2845, 1320, 4997, 638, 3191, 1464, 601, 1223, 2791, 3144,
3678, 3659, 650, 2810, 3403, 3540, 2984, 1685, 2817, 4879, 1943,
3412, 4797, 1695, 554, 2866, 813, 3749, 367, 1325, 1638, 1494,
1970, 1509, 42, 2060, 1381, 1760, 4188, 619, 172, 331, 3695, 1399,
3172, 946, 3794, 4975, 1175, 3756, 3687, 1383, 2917, 3688, 4758,
714, 2665, 566, 226, 4821, 4661, 1645, 3663, 4370, 1642, 3498,
3055, 3543, 2606, 3203, 2136, 4858, 989, 1969, 4098, 4362, 2895,
1713, 2278, 1352, 4967, 3934, 1789, 3929, 2532, 2941, 1807, 4933,
2830, 3536, 2859, 3185, 4783, 1395, 402, 2964, 2715, 2117, 1892,
3175, 3503, 3522, 1006, 2581, 4912, 3889, 2334, 4482, 3186, 2422,
2106, 1487, 1237, 950, 1773, 1157, 2870, 3070, 4806, 4825, 3149,
4882, 1179, 3857, 3210, 4991, 844, 2153, 2858, 1061, 4378, 642,
4574, 4533, 471, 2766, 1723, 4900, 786, 746, 3501, 1362, 4893,
4376, 4107, 858, 3825, 1598, 4911, 2088, 2955, 3285, 308, 1465,
1840, 4090, 4270, 485, 4192, 4265, 4598, 242, 3747, 3572, 2772,
2103, 1756, 3923, 2398, 4604, 1143, 2785, 3097, 2860,
854, 1266, 1421, 1423, 1573, 2399, 4253, 4423, 4464
\end{quote}

Our distilled $2$B model is additionally evaluated on the full
$5{,}000$-stream test set (TABME-full); the subset / full numbers
track closely (DocF1 $0.756$ vs.\ $0.770$ for
Qwen-3.5-2B$^{\star,\mathrm{PwT}}$), supporting the $250$-stream
subset as representative of the full distribution.

\section{TABME Page-Level Precision/Recall}
\label{app:tabme-pr}

Table~\ref{tab:tabme-pr} reports page-level precision and recall on
TABME-250 under RICoT prompting. Recall is $\geq 0.955$ on every
prompted model and precision spans $0.733$--$0.873$, so errors are
mostly extra boundaries rather than missed ones. The distilled
student shows the same shape on the full test set (boundary-level
precision $0.836$, recall $0.951$, B-F1 $0.890$).
Appendix~\ref{app:tabme-errors} analyses these extra boundaries:
some are document starts that TABME's file-level labels merge, but
most are errors of the model.

\begin{table}[h]
  \centering
  \small
  \setlength{\tabcolsep}{6pt}
    \begin{tabular}{lccc}
    \toprule
    Model & PageP & PageR & Gap \\
    \midrule
    G4.7-F    & 0.733 & 0.955 & $+0.222$ \\
    G4.7      & 0.832 & 0.962 & $+0.130$ \\
    C-H4.5    & 0.861 & 0.965 & $+0.104$ \\
    G5        & 0.873 & 0.978 & $+0.105$ \\
    \bottomrule
  \end{tabular}
\caption{TABME-250 page-level precision/recall under RICoT
  prompting.}
  \label{tab:tabme-pr}
\end{table}

\section{TABME False Boundaries: Error Analysis}
\label{app:tabme-errors}

\paragraph{Setup.} We analyse the false boundaries (FPs) of
Q3.5-2B$^{\star}$ (RICoT, no pagewise thinking) on the full
$5{,}000$-stream test set, scoring sub-document starts as
continuations as in all our metrics (Section~\ref{sec:metrics}):
precision $0.836$, recall $0.951$, B-F1 $0.890$, $9{,}331$ FPs. We
sampled $100$ FPs: $45$ at random (one per stream) and $55$ as one
error site per source file, from files not already sampled; they
cover $98$ distinct (file, page) sites. Each was judged by hand
from the OCR text, and $25$ of them, including every A1 site, were
also checked against the page images; this changed one label (a
magazine letters page first judged to come from another issue was
the next page of the same issue).

\paragraph{Categories.}
\textbf{A1}: a genuine top-level document begins inside one source
file (e.g.\ two separate funding records filed together); TABME's
file-level label merges them.
\textbf{A2}: a genuine nested document (envelope, enclosure,
attachment, fax payload) that the model emitted as a top-level
start. In our framework the correct output is a sub-document start,
which the metric scores as continuation, so A2 is a model error:
the right document at the wrong level.
\textbf{B}: no document starts here; the page continues the same
document.
\textbf{C}: ambiguous (archive slip sheets, near-empty pages,
unreadable OCR).

\begin{table}[h]
  \centering
  \small
  \setlength{\tabcolsep}{4pt}
  \begin{tabular}{lcccc}
    \toprule
    Weighting & A1 & A2 & B & C \\
    \midrule
    Per sampled case ($n{=}100$)      & 14 & 27 & 44 & 15 \\
    Per distinct site ($98$)          & 13 & 27 & 45 & 15 \\
    Per FP occurrence                 & 17 & 33 & 33 & 17 \\
    \bottomrule
  \end{tabular}
\caption{Share (\%) of TABME false boundaries by category,
  Q3.5-2B$^{\star}$, full test set. ``Per FP occurrence'' weights
  each site by how often it is a false boundary across the test set.
  Each share is uncertain by about $\pm10$ points.}
  \label{tab:tabme-fp-cats}
\end{table}

\paragraph{Findings.} Table~\ref{tab:tabme-fp-cats} gives the breakdown. A1 and A2 sites recur more often than B sites
($\sim$$17\times$ vs.\ $\sim$$9.6\times$ across the test set): short
files such as a letter and its envelope, or a memo and its
attachment, appear in many streams. Per occurrence, A1 accounts for
about $17\%$ of the FPs; counting these pages as correct raises
precision from $0.836$ to about $0.86$. A1 also includes convention
cases, such as a second copy of a form or a technical sheet followed
by its safety data sheet. About two thirds of the FPs (A2 and B) are
errors of the model. The B errors follow a few patterns:
(i)~later pages that repeat the title block, project or estimate
number, or carry ``Page~$n$ of~$m$''; (ii)~section, table and figure
headings inside a long document; (iii)~title slides or cover pages
inside a deck or brochure; (iv)~back sides and stamp pages (e.g.\
the endorsement side of a check); and (v)~a memo that follows its
own distribution page.

\begin{figure*}[t]
  \centering
  \begin{minipage}[t]{0.48\textwidth}\centering
    \includegraphics[width=\linewidth]{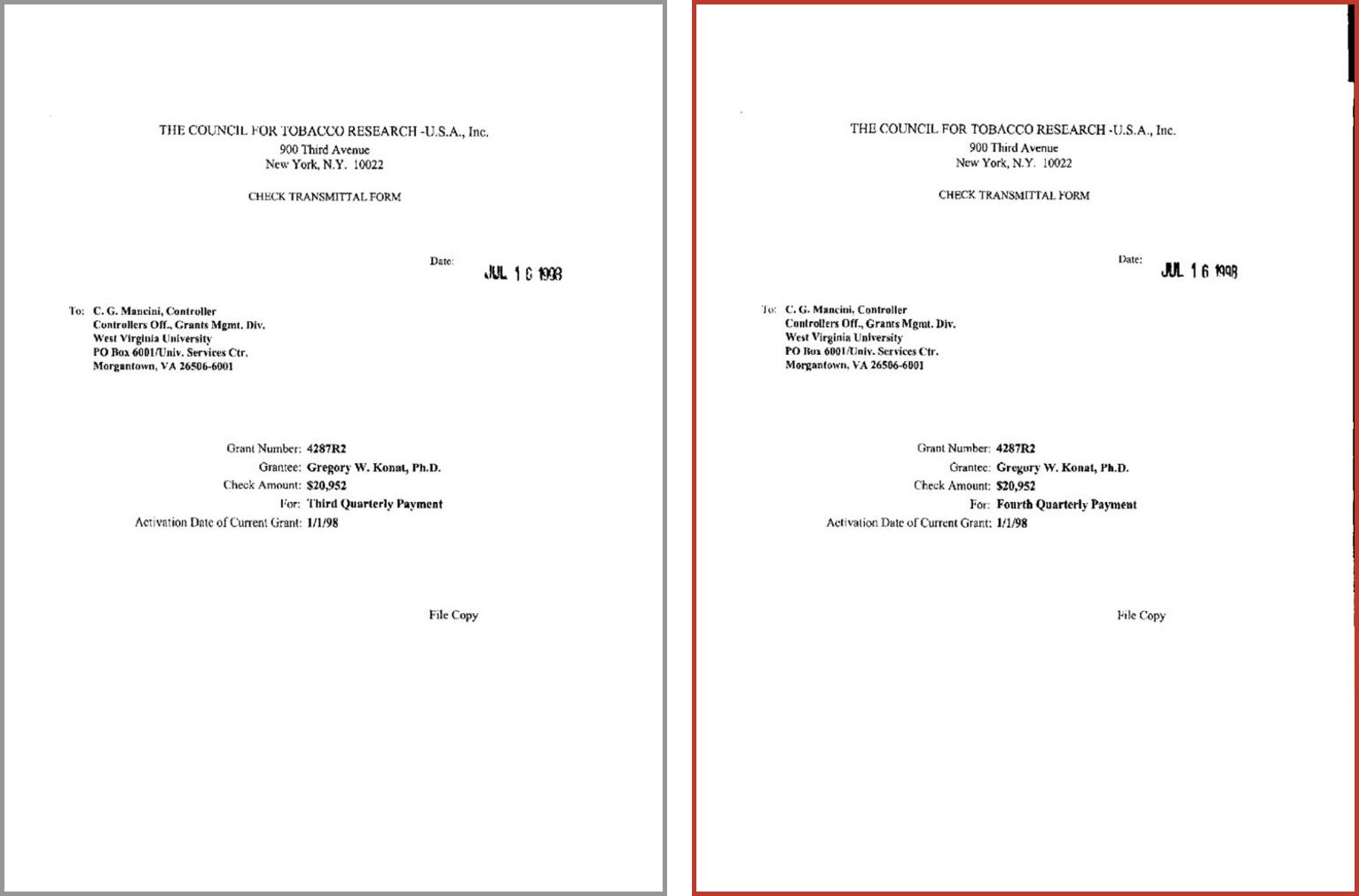}\\[1pt]
    {\small (a) Check transmittal forms, third and fourth payment}
  \end{minipage}\hfill
  \begin{minipage}[t]{0.48\textwidth}\centering
    \includegraphics[width=\linewidth]{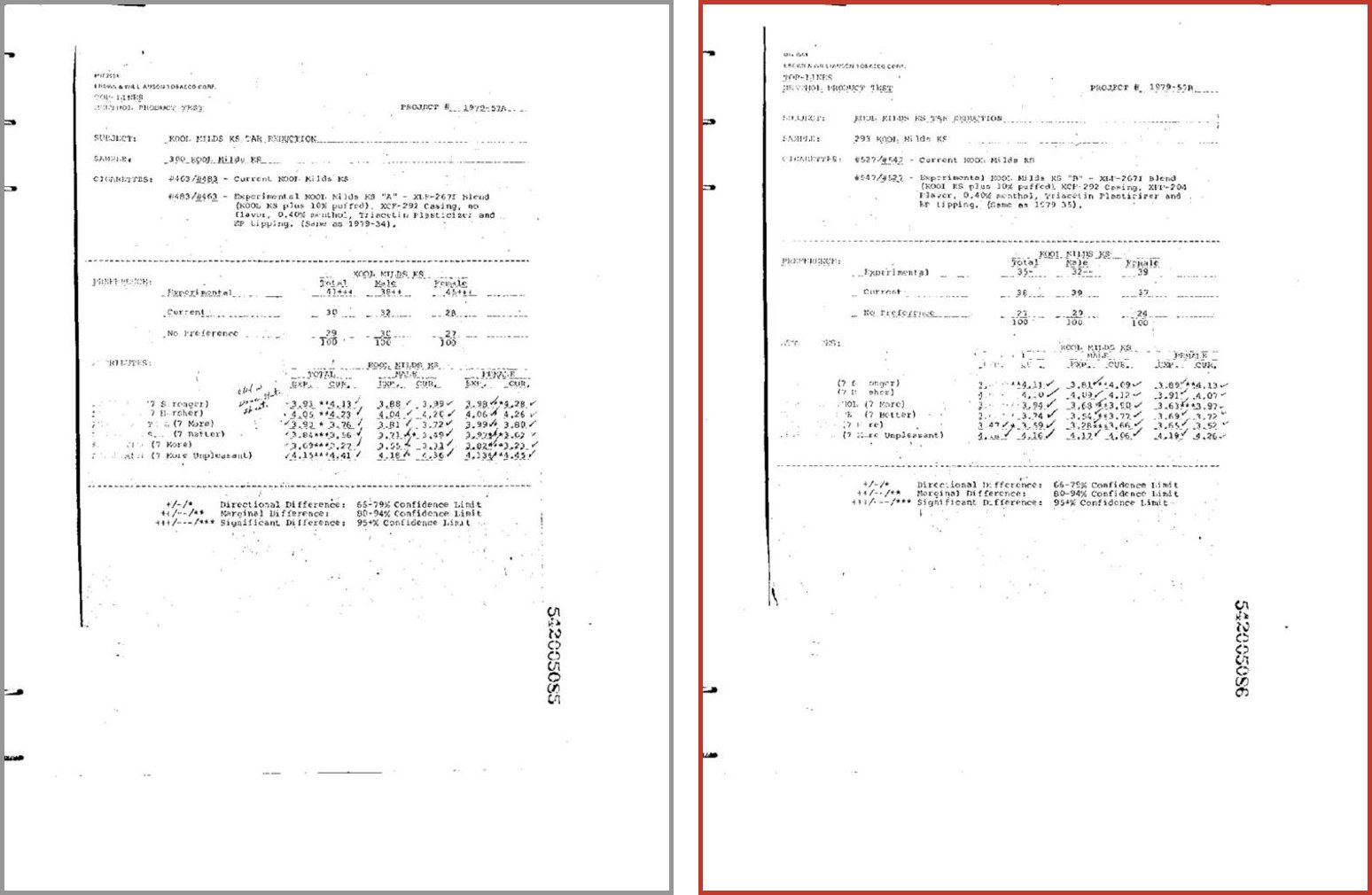}\\[1pt]
    {\small (b) Product-test summaries of two projects}
  \end{minipage}\\[6pt]
  \begin{minipage}[t]{0.48\textwidth}\centering
    \includegraphics[width=\linewidth]{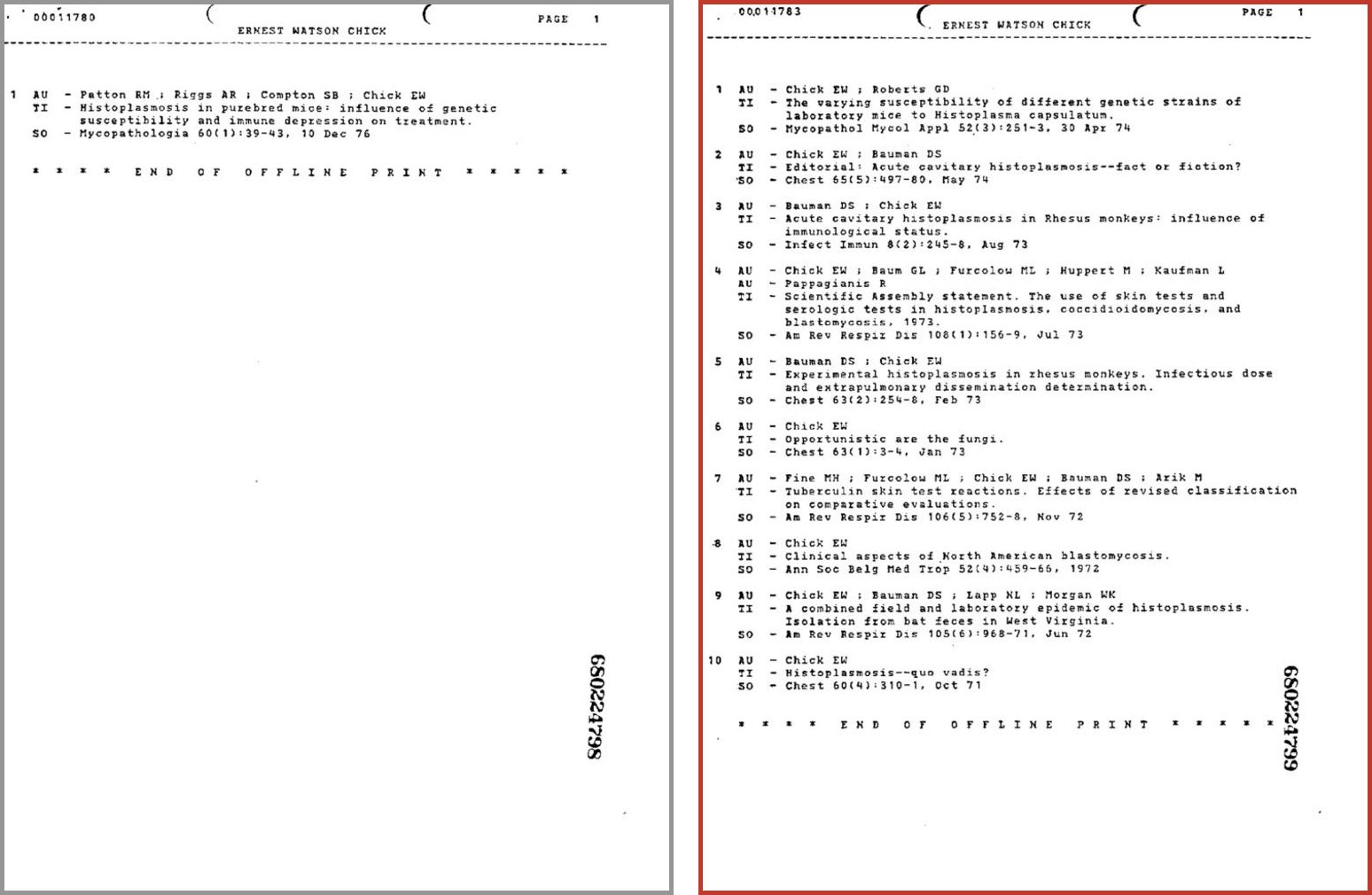}\\[1pt]
    {\small (c) Two separate database printouts}
  \end{minipage}\hfill
  \begin{minipage}[t]{0.48\textwidth}\centering
    \includegraphics[width=\linewidth]{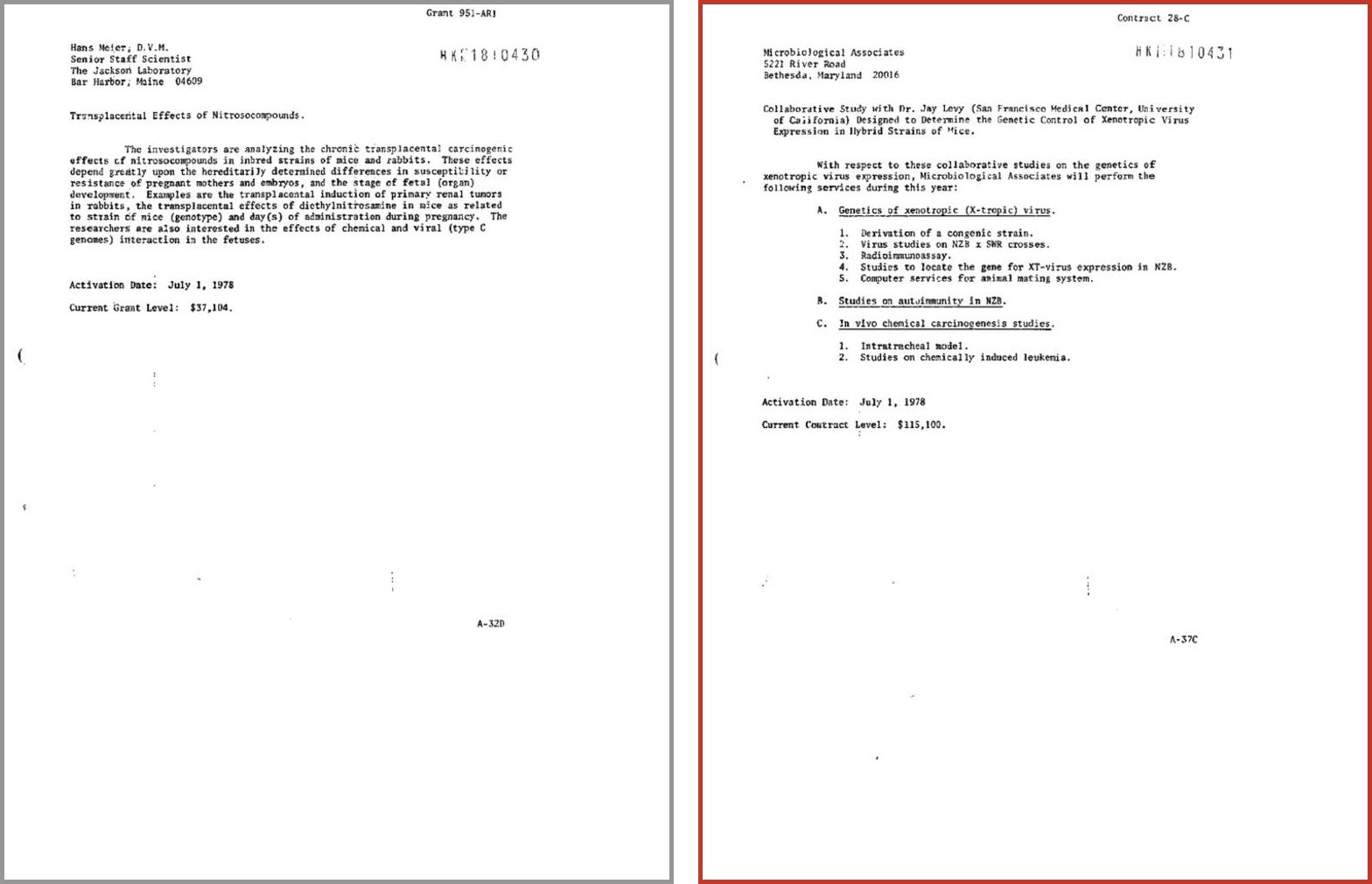}\\[1pt]
    {\small (d) A grant record followed by a contract record}
  \end{minipage}
  \caption{Separate documents inside a single TABME source file. In
    each pair, the right page (red border) is where the model started
    a new document; TABME's ground truth treats it as a continuation
    of the left page (grey border) because both belong to the same
    source PDF.}
  \label{fig:tabme-fp}
\end{figure*}

\paragraph{Separate documents within one source file.}
Figure~\ref{fig:tabme-fp} shows four A1 cases, in which the page
the model marked as a new document begins a separate document filed
in the same source PDF; TABME's file-level label counts each
boundary as an error. (a)~Two check transmittal forms of the
Council for Tobacco Research for the same grant, one for the third
and one for the fourth quarterly payment (stream 0390, p.~11; a
false boundary in $18$ of the $33$ streams containing the file).
(b)~Product-test summaries of two different projects, with
different samples and results (stream 3712, p.~7; $27$ of $29$).
(c)~Two separate bibliographic database printouts, each with its
own print number, ``PAGE 1'' header and end-of-print marker (stream
0065, p.~4; $11$ of $32$). (d)~Compiled funding records: a grant
followed by a contract to a different recipient, with its own
activation date and amount (stream 0211, p.~14; $28$ of $28$).

\section{Token-Cost Breakdown}
\label{app:token-cost}

Table~\ref{tab:token-cost} expands the cost figures in
Section~\ref{sec:thinking-ablation} with per-channel breakdowns,
all on GLM-4.7-flash with FinancePSS-205 input. Input is
approximately constant across modes; output variation is dominated
by the internal-thinking budget, which constitutes $73\%$--$82\%$
of total output in internal modes. When internal thinking is on,
external pagewise-thinking tokens drop from $72$ to $44$
($\sim$$39\%$ reduction), evidence of partial substitution between
the two reasoning channels.

\begin{table}[h]
  \centering
  \footnotesize
  \setlength{\tabcolsep}{3pt}
    \begin{tabular}{@{}lrrrrrr@{}}
    \toprule
    Mode & In & Out & Int. & Ext. & Dec. & C-fwd \\
    \midrule
    no-think     & 752 & 172 & 0   & 0  & 142 & 21 \\
    pagewise     & 797 & 232 & 0   & 72 & 126 & 20 \\
    windowed     & 772 & 251 & 0   & 83 & 128 & 20 \\
    int.+pagewise & 767 & 722 & 528 & 44 & 115 & 16 \\
    int. only    & 736 & 823 & 674 & 0  & 110 & 17 \\
    \bottomrule
  \end{tabular}
\caption{Per-page output-token cost on GLM-4.7-flash,
  FinancePSS-205, broken out by component. Input is roughly
  constant across modes. \emph{Int.} = internal reasoning tokens;
  \emph{Ext.} = external (pagewise) reasoning tokens.}
  \label{tab:token-cost}
\end{table}

\section{Interleaved-Thinking Ablation (Full)}
\label{app:abl-thinking}

Table~\ref{tab:abl-thinking} expands the pagewise-thinking ablation
across all three datasets and all three prompted models. The
\emph{think} row is the full RICoT configuration with pagewise
thinking; \emph{windowed\_think} replaces the interleaved per-page
reasoning with a single block of reasoning emitted before the
window-level decisions; \emph{wo\_think} disables structured
reasoning entirely. FinancePSS uses with-expand.

\begin{table}[h]
  \centering
  \footnotesize
  \setlength{\tabcolsep}{4pt}
    \begin{tabular}{llccc}
    \toprule
    Dataset & Configuration & G4.7-F & G4.7 & G5 \\
    \midrule
    \multirow{3}{*}{FinancePSS}
      & think (full)     & 0.762 & 0.832 & 0.826 \\
      & windowed\_think  & 0.754 & 0.814 & 0.746 \\
      & wo\_think        & 0.688 & 0.784 & 0.738 \\
    \midrule
    \multirow{3}{*}{Tobacco800}
      & think (full)     & 0.867 & 0.961 & 0.967 \\
      & windowed\_think  & 0.846 & 0.968 & 0.981 \\
      & wo\_think        & 0.822 & 0.942 & 0.981 \\
    \midrule
    \multirow{3}{*}{TABME-250}
      & think (full)     & 0.683 & 0.794 & 0.847 \\
      & windowed\_think  & 0.677 & 0.805 & 0.847 \\
      & wo\_think        & 0.661 & 0.762 & 0.829 \\
    \bottomrule
  \end{tabular}
\caption{Interleaved-thinking ablation, DocF1 by dataset and
  model.}
  \label{tab:abl-thinking}
\end{table}

\section{State-Propagation Ablation (Full)}
\label{app:abl-state}

Table~\ref{tab:abl-state} expands the state-propagation ablation
across all three datasets and three models. The full RICoT
configuration carries both the reasoning history $\mathcal{H}^{(c)}$
(``cognitive trace'') and the active-segment summary
$\mathcal{A}^{(c)}$ (which includes the model-emitted
\texttt{active\_segment\_carry\_forward} block) across windows.
\emph{only\_cognitive\_trace} keeps only $\mathcal{H}^{(c)}$;
\emph{only\_active\_segment} keeps only $\mathcal{A}^{(c)}$;
\emph{no\_carry\_forward} drops the model-emitted carry-forward
block but keeps the deterministic summary;
\emph{no\_context\_propagation} disables all cross-window state.

\begin{table}[h]
  \centering
  \footnotesize
  \setlength{\tabcolsep}{4pt}
    \begin{adjustbox}{max width=\linewidth}
\begin{tabular}{llccc}
    \toprule
    Dataset & Configuration & G4.7-F & G4.7 & G5 \\
    \midrule
    \multirow{5}{*}{FinancePSS}
      & full                    & 0.762 & 0.832 & 0.826 \\
      & only\_cognitive\_trace  & 0.748 & 0.824 & 0.766 \\
      & only\_active\_segment   & 0.749 & 0.801 & 0.786 \\
      & no\_carry\_forward      & 0.736 & 0.841 & 0.795 \\
      & no\_context\_prop.      & 0.752 & 0.837 & 0.778 \\
    \midrule
    \multirow{5}{*}{Tobacco800}
      & full                    & 0.867 & 0.961 & 0.967 \\
      & only\_cognitive\_trace  & 0.887 & 0.971 & 0.981 \\
      & only\_active\_segment   & 0.892 & 0.935 & 0.968 \\
      & no\_carry\_forward      & 0.910 & 0.968 & 0.967 \\
      & no\_context\_prop.      & 0.892 & 0.961 & 0.977 \\
    \midrule
    \multirow{5}{*}{TABME-250}
      & full                    & 0.683 & 0.794 & 0.847 \\
      & only\_cognitive\_trace  & 0.692 & 0.799 & 0.854 \\
      & only\_active\_segment   & 0.668 & 0.789 & 0.818 \\
      & no\_carry\_forward      & 0.677 & 0.816 & 0.855 \\
      & no\_context\_prop.      & 0.670 & 0.751 & 0.796 \\
    \bottomrule
  \end{tabular}
  \end{adjustbox}
\caption{State-propagation ablation, DocF1.}
  \label{tab:abl-state}
\end{table}

The cognitive trace is consistently load-bearing (removing it
hurts noticeably), while the carry-forward block is mixed
across models---supporting the framing in
Section~\ref{sec:results-ablations} that the state-propagation
story rests on the cognitive trace rather than carry-forward
specifically.

\section{Decision-Block Ablation (Full)}
\label{app:abl-decision}

Table~\ref{tab:abl-decision} ablates the per-page decision schema.
\emph{barebone} drops everything except $y_i$ and $\ell_i$;
\emph{no\_auxiliary} drops auxiliary tags
($\langle\text{page\_role}\rangle$,
$\langle\text{sub\_document\_start}\rangle$);
\emph{no\_grounding} drops the fingerprint and evidence quote;
\emph{no\_nested} drops the sub-document axis $z_i$. The pattern
inverts with model size: smaller models often pay a capacity tax
for schema overhead, while larger models are roughly invariant or
benefit from the structure.

\begin{table}[h]
  \centering
  \footnotesize
  \setlength{\tabcolsep}{4pt}
    \begin{tabular}{llccc}
    \toprule
    Dataset & Configuration & G4.7-F & G4.7 & G5 \\
    \midrule
    \multirow{5}{*}{FinancePSS}
      & full               & 0.762 & 0.832 & 0.826 \\
      & barebone           & 0.708 & 0.772 & 0.760 \\
      & no\_auxiliary      & 0.775 & 0.809 & 0.770 \\
      & no\_grounding      & 0.695 & 0.831 & 0.755 \\
      & no\_nested         & 0.740 & 0.812 & 0.746 \\
    \midrule
    \multirow{5}{*}{Tobacco800}
      & full               & 0.867 & 0.961 & 0.967 \\
      & barebone           & 0.939 & 0.961 & 0.987 \\
      & no\_auxiliary      & 0.910 & 0.971 & 0.971 \\
      & no\_grounding      & 0.851 & 0.967 & 0.977 \\
      & no\_nested         & 0.824 & 0.968 & 0.977 \\
    \midrule
    \multirow{5}{*}{TABME-250}
      & full               & 0.683 & 0.794 & 0.847 \\
      & barebone           & 0.661 & 0.812 & 0.851 \\
      & no\_auxiliary      & 0.672 & 0.804 & 0.849 \\
      & no\_grounding      & 0.647 & 0.791 & 0.841 \\
      & no\_nested         & 0.656 & 0.802 & 0.845 \\
    \bottomrule
  \end{tabular}
\caption{Decision-block ablation, DocF1.}
  \label{tab:abl-decision}
\end{table}

\paragraph{Class labels.} Table~\ref{tab:abl-decision} does not
isolate the class labels, so we ablated them separately on
GLM-4.7-flash / FinancePSS, $5$ runs each, removing each label from
prompt, output, and parser (Table~\ref{tab:abl-class}). The
top-level class $\ell_i$ is never scored; the sub-document class is
a direct input to the with-expand protocol, which promotes a nested
start only when its predicted class is in the priority set.
Removing both labels changes DocF1 by $-0.011$, within noise, so
the top-level label is metric-neutral; we keep it as a reasoning
scaffold (naming the active document anchors the continuation test)
and because it is the field downstream routing consumes. The
sub-document label is what lifts the full model from $0.739$
(no-expand) to $0.753$ (with expand); removing it removes that gain
by construction. For pure segment boundaries the nested axis can be
dropped at no cost: removing only the sub-document class is neutral
on F1 and lowers $\mathrm{MNDD}_\mu$.

\begin{table}[h]
  \centering
  \scriptsize
  \setlength{\tabcolsep}{2.5pt}
  \begin{adjustbox}{max width=\columnwidth}
  \begin{tabular}{@{}lccc@{}}
    \toprule
    Configuration & DocF1 & PageF1 & $\mathrm{MNDD}_\mu$ \\
    \midrule
    Full, with expand        & $0.753{\pm}0.024$ & $0.892{\pm}0.016$ & $1.14{\pm}0.23$ \\
    Full, no expand          & $0.739{\pm}0.013$ & $0.878{\pm}0.013$ & $1.32{\pm}0.19$ \\
    $-$ sub-document class   & $0.743{\pm}0.021$ & $0.884{\pm}0.015$ & $1.08{\pm}0.14$ \\
    $-$ both class labels    & $0.728{\pm}0.027$ & $0.872{\pm}0.015$ & $1.40{\pm}0.32$ \\
    \bottomrule
  \end{tabular}
  \end{adjustbox}
\caption{Class-label ablation, GLM-4.7-flash, FinancePSS-205,
  $5$-run mean $\pm$ $95\%$ CI. Without the sub-document class,
  with- and no-expand scores coincide.}
  \label{tab:abl-class}
\end{table}

\section{Per-Dataset Full Result Dumps}
\label{app:full-results}

Complete per-dataset result tables---all metrics
(DocF1, PageF1, PageP, PageR, $\mathrm{MNDD}_\mu$, STP\%); all
backbone${\times}$method${\times}$ablation combinations;
both with-expand and without-expand variants on FinancePSS---will be
included in the public release.

\section{Training-Sample Configuration}
\label{app:sampling-config}

Table~\ref{tab:variant-probs} lists the probabilities used when
rendering training samples (Section~\ref{sec:training-modes}): the
chunk-level probability of omitting page-level reasoning (first row)
and the per-field probabilities of omitting optional decision and
carry-forward fields. The two are sampled independently. Values were
set heuristically, weighted by field importance, and not subjected
to hyperparameter sweeps owing to compute constraints.

\begin{table}[h]
  \centering
  \small
  \setlength{\tabcolsep}{6pt}
    \begin{adjustbox}{max width=\linewidth}
\begin{tabular}{lc}
    \toprule
    Omitted content & Probability \\
    \midrule
    Pagewise reasoning blocks (chunk-level) & 0.40 \\
    Per-page: \texttt{document\_class} & 0.20 \\
    Per-page: \texttt{confidence}      & 0.20 \\
    Per-page: \texttt{grounding} (fingerprint+evidence) & 0.20 \\
    Per-page: \texttt{sub\_document} nested instance & 0.15 \\
    Carry-forward: \texttt{progress} field & 0.30 \\
    \bottomrule
  \end{tabular}
  \end{adjustbox}
\caption{Probabilities of omitting content when rendering training
  samples (Section~\ref{sec:training-modes}).}
  \label{tab:variant-probs}
\end{table}

\noindent\emph{Note.} The probabilities above are working values
used in the single training run reported in the paper. We did not
sweep them; their effect on accuracy remains unexplored.

\section{Training Corpus Attrition Breakdown}
\label{app:attrition}

Table~\ref{tab:training-pipeline} summarizes the per-stage attrition
of the FinePDFs-derived training corpus. Of the $300$K candidate
documents sampled from the $12.3$M-document finance subset, roughly
$50\%$ survive download and language-alignment filtering. The main
attrition stages are: failed Common Crawl fetch (cache miss / dead
URL), page-count mismatch between the downloaded PDF and FinePDFs
metadata (indicating unreliable OCR alignment), and detection of
pages in scripts unsupported by our OCR model. The surviving
$150{,}448$ documents feed into teacher labeling.

\begin{table}[h]
  \centering
  \small
  \setlength{\tabcolsep}{6pt}
    \begin{adjustbox}{max width=\linewidth}
\begin{tabular}{lr}
    \toprule
    Stage & Count \\
    \midrule
    FinePDFs (English subset) & $\sim$206M docs \\
    Finance filter (stages 1+2) & 12.3M docs \\
    Sampled for training pool & 300K docs \\
    Download, re-OCR, language filter & 150{,}448 docs \\
    Teacher labeling (chunk samples) & 467{,}229 chunks \\
    Final training corpus & 46{,}791 docs \\
    \quad multi-segment & 20{,}959 \\
    \quad single-segment & 25{,}832 \\
    \bottomrule
  \end{tabular}
  \end{adjustbox}
\caption{Training corpus construction pipeline.}
  \label{tab:training-pipeline}
\end{table}

\section{Training Hyperparameters}
\label{app:hparams}

Full QLoRA hyperparameters used for Qwen-3.5-2B fine-tuning
(Section~\ref{sec:training-setup}):

\begin{itemize}
\setlength{\itemsep}{0pt}
\item LoRA rank $32$; $\alpha=64$; dropout $0$; target modules
\texttt{q\_proj, k\_proj, v\_proj, o\_proj, gate\_proj, up\_proj,
down\_proj}.
\item Optimizer: AdamW-$8$bit; weight decay $0.01$; max grad norm
$1.0$.
\item Learning rate: $8{\times}10^{-5}$ with cosine schedule and
$8\%$ warmup ratio.
\item Batch size: per-device $8$, gradient-accumulation $2$
(effective $16$); evaluation batch size $32$.
\item Epochs: $1$; validation split $1\%$ ($\sim$$1{,}800$ samples).
\item Precision: bf16; gradient checkpointing on.
\item Hardware: $4{\times}$NVIDIA A100 80GB; wall-clock $\sim$$44$
hours.
\end{itemize}

\section{Worked Training-Sample Example}
\label{app:samples}

A complete example showing the same source chunk rendered into a
training sample (i)~with all schema fields present and pagewise
reasoning emitted, (ii)~without page-level reasoning (decisions
only), and (iii)~decisions only, with several optional decision
fields also omitted, will be included in the public release.

\section{FinePDFs Finance-Filter Classifier}
\label{app:finepdf}

The Stage~1 finance lexicon contains three tiers of keyword
patterns of increasing specificity (high-confidence finance terms,
moderate-confidence terms, and disambiguation terms used to
distinguish between finance vs.\ accounting vs.\ marketing
contexts). The Stage~2 XGBoost classifier was trained on
$\sim$$100$K Stage-1-surviving documents labeled by an LLM judge
(prompted to apply binary finance vs.\ non-finance labels using a
detailed rubric); we used TF-IDF features with a $50$K-term
vocabulary capped at unigrams and bigrams. The decision threshold
was tuned on a $5$K-document held-out set to maximize F$_{0.5}$
(precision-weighted). The lexicon files, the LLM-judge prompt, and
the fitted TF-IDF vectorizer will be included in the public
release.

\section{Prompt Templates and Inference Configuration}
\label{app:prompts}

Structural prompt templates for RICoT and both baselines (task
framing, stream-context format, and per-page decision schema) will be
included in the public release; the verbatim instruction
prose is withheld. All inference runs use
temperature $0.001$ and top-$p$ $1$; other sampling parameters are left
at each provider's defaults. Internal extended thinking is
explicitly disabled in the RICoT runs and enabled in the baseline
runs. Residual provider-side nondeterminism is quantified in
Appendix~\ref{app:uncertainty}.

\section{Pseudocode for the RICoT Inference Loop}
\label{app:algorithm}

Algorithm~\ref{alg:ricot} gives the full inference-time procedure
referenced in Section~\ref{sec:ricot-interleaved}.

\begin{algorithm}[H]
\small
\begin{algorithmic}[1]
\REQUIRE Stream $\mathbf{X}$, token budget $B$, page cap $w_{\max}$, overlap $o\!=\!1$, history size $k\!=\!6$
\ENSURE Decisions $\mathbf{d}=(d_1,\dots,d_N)$
\STATE $\mathcal{H}\!\gets\!\emptyset$;\ $\mathcal{A}\!\gets\!\emptyset$;\ $\kappa\!\gets\!\emptyset$;\ $\mathcal{C}\!\gets\!\emptyset$;\ $c\!\gets\!1$
\WHILE{pages remain}
  \STATE $W_c\!\gets\textsc{PackWindow}(\mathbf{X};\,B,w_{\max},o)$ \hfill // greedy until budget or cap
  \STATE tag $W_c$ with \texttt{<stream>}/\texttt{</stream>} if first/last
  \STATE $(\mathbf{T}_c,\mathbf{d}_c,\kappa') \gets \textsc{ICoT}(W_c,\mathcal{H},\mathcal{A}\cup\kappa)$
  \STATE $\mathbf{d}_c \gets \textsc{Validate}(\mathbf{d}_c,\mathcal{C})$ \hfill // cache-fill, retry, fallback
  \STATE $\mathcal{C}\!\gets\!\mathcal{C}\cup\{d:\text{valid}\}$;\ append $\mathbf{d}_c$ to $\mathbf{d}$
  \STATE $\mathcal{H}\!\gets$ last $k$ pairs of $(\mathbf{T}_c,\mathbf{d}_c)$
  \IF{$\exists\,d\in\mathbf{d}_c:y\!=\!\text{S}$} \STATE $\mathcal{A}\!\gets$ metadata from new segment
  \ELSE \STATE $\mathcal{A}\!\gets\!\mathcal{A}\cup\mathbf{d}_c$
  \ENDIF
  \STATE $\kappa\!\gets\!\kappa'$ (\,$\emptyset$ if final window)
  \STATE $c\!\gets\!c+1$
\ENDWHILE
\RETURN $\mathbf{d}$
\end{algorithmic}
\caption{RICoT page-stream segmentation}
\label{alg:ricot}
\end{algorithm}

\end{document}